\documentclass[10pt,twocolumn,letterpaper]{article}

\usepackage{cvpr}              

\usepackage{float}
\usepackage[algo2e]{algorithm2e}

\definecolor{cvprblue}{rgb}{0.21,0.49,0.74}
\usepackage[pagebackref,breaklinks,colorlinks,allcolors=cvprblue]{hyperref}

\usepackage{graphicx}
\usepackage{booktabs}
\usepackage{bbm}
\usepackage{makecell}
\usepackage{bm}
\usepackage{multicol}
\usepackage{multirow}
\usepackage{caption}
\usepackage{xcolor}
\usepackage{bbding}
\usepackage{pifont}
\usepackage{wrapfig}

\usepackage{url}
\usepackage{color}
\usepackage{graphicx}
\usepackage{multirow}
\usepackage{array}
\usepackage{booktabs}
\usepackage{ulem}
\usepackage{enumitem}
\usepackage{algpseudocode}
\usepackage{algorithmicx,algorithm}

\long\def\comment#1{}

\makeatletter
\newcommand{\printfnsymbol}[1]{%
  \textsuperscript{\@fnsymbol{#1}}%
}
\makeatother

\def\paperID{12618} 
\def\confName{CVPR}
\def\confYear{2025}

\title{Protecting Your Video Content: Disrupting Automated Video-based LLM Annotations}

\author{Haitong Liu\textsuperscript{\rm 1}\thanks{Equal contribution.}, \ Kuofeng Gao\textsuperscript{\rm 1}\printfnsymbol{1}, \ Yang Bai\textsuperscript{\rm 2}\thanks{Corresponding authors.} , \ Jinmin Li\textsuperscript{\rm 1}, \  Jinxiao Shan\textsuperscript{\rm 3}, \ Tao Dai\textsuperscript{\rm 4}\printfnsymbol{2}, \ Shu-Tao Xia\textsuperscript{\rm 1,\rm 5}\\
\textsuperscript{\rm 1} Tsinghua University \quad
\textsuperscript{\rm 2} ByteDance \quad
\textsuperscript{\rm 3} Digital Center, China Merchants Group Limited \\
\textsuperscript{\rm 4} Shenzhen University \quad
\textsuperscript{\rm 5} Peng Cheng Laboratory
\\
\tt\small \{liuhaito24,gkf21,ljm22\}@mails.tsinghua.edu.cn, baiyang0522@gmail.com \\
\tt\small  shanjinxiao@cmhk.com, daitao.edu@gmail.com, xiast@sz.tsinghua.edu.cn
}

\begin{document}
\maketitle
\begin{abstract}
Recently, video-based large language models (video-based LLMs) have achieved impressive performance across various video comprehension tasks. However, this rapid advancement raises significant privacy and security concerns, particularly regarding the unauthorized use of personal video data in automated annotation by video-based LLMs. These unauthorized annotated video-text pairs can then be used to improve the performance of downstream tasks, such as text-to-video generation. To safeguard personal videos from unauthorized use, we propose two series of protective video watermarks with imperceptible adversarial perturbations, named \textbf{Ramblings} and \textbf{Mutes}. Concretely, \textbf{Ramblings} aim to mislead video-based LLMs into generating inaccurate captions for the videos, thereby degrading the quality of video annotations through inconsistencies between video content and captions. \textbf{Mutes}, on the other hand, are designed to prompt video-based LLMs to produce exceptionally brief captions, lacking descriptive detail. Extensive experiments demonstrate that our video watermarking methods effectively protect video data by significantly reducing video annotation performance across various video-based LLMs, showcasing both stealthiness and robustness in protecting personal video content. Our code is available at \url{https://github.com/ttthhl/Protecting_Your_Video_Content}.
\end{abstract}    
\section{Introduction}
\label{sec:intro}

Recent years have witnessed significant advancements in both video-based large language models (video-based LLMs)~\cite{maaz2023video,tan2024koala,xu2024pllava,yang2023vid2seq,zhang2023video,huang2024vtimellm} and text-to-video generation models~\cite{chen2023videocrafter1,guo2023animatediff,lin2024animatediff,ma2024latte,wang2024animatelcm}. These two types of models demonstrate a close relationship: text-to-video models rely heavily on high-quality video-text pairs for fine-tuning, while video-based LLMs can enhance video-text pairs quality through advanced annotation capabilities.
By leveraging the power of large language models~\cite{brown2020language,touvron2023llama,wei2022chain,gao2024embedding,gao2024denial}, video-based LLMs have revolutionized video content interpretation and annotation. This advancement directly contributes to the improvement of video data annotation quality.
However, alongside these technological advances, concerns about data safety and privacy~\cite{li2022untargeted,bai2022practical,li2022move,li2023black,li2024video,deng2024deconstructing,wei2024pointncbw} have emerged. Given the vast amount of personal video content available on multimedia platforms, protecting this data from unauthorized annotation has become an urgent challenge that requires immediate attention.
\begin{figure}
    \centering
    \includegraphics[trim={10 110 0 0},clip,width=0.95\linewidth]{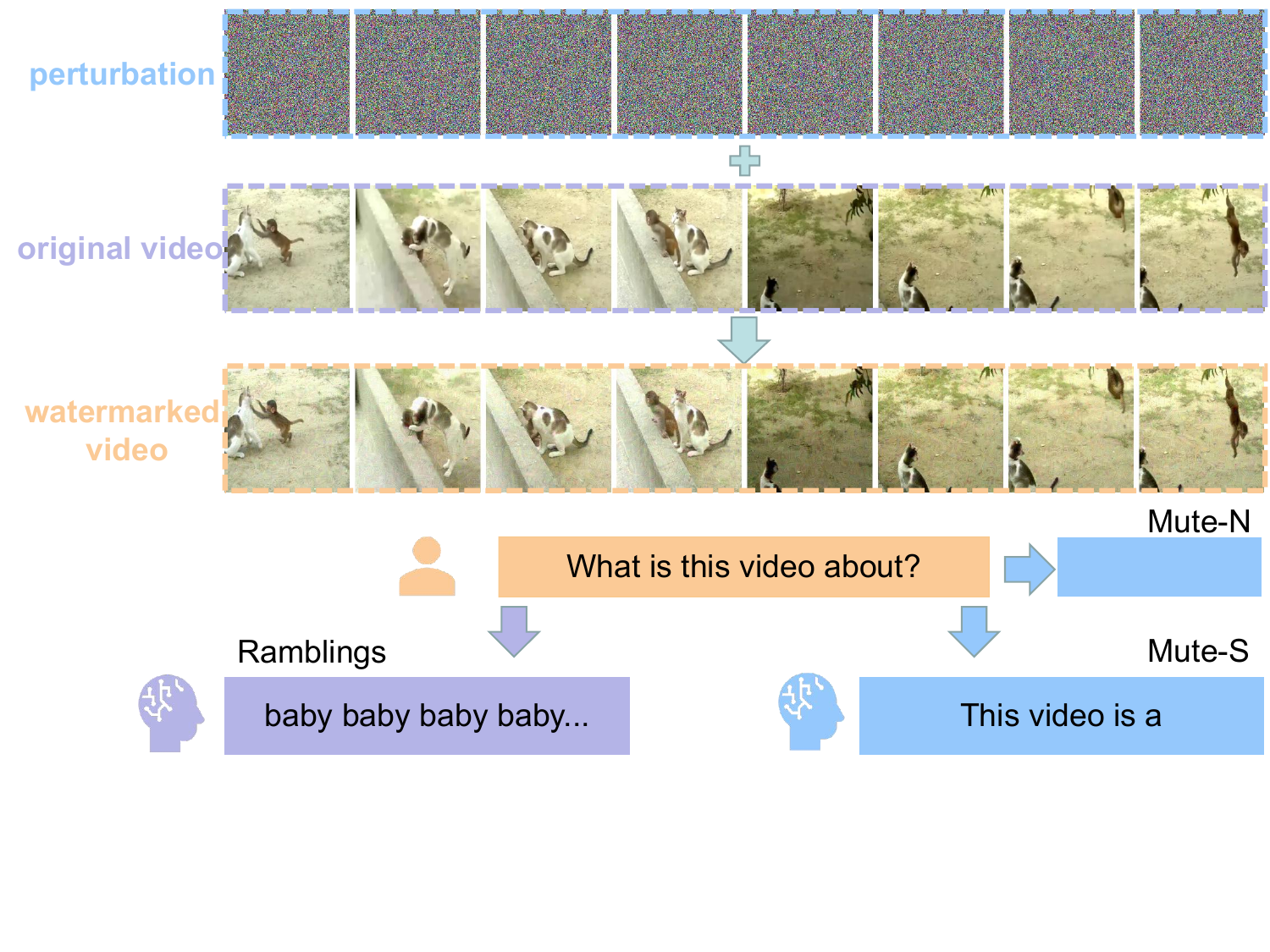}
    \caption{We propose two strategies to protect the video content: Ramblings and Mutes. The Ramblings approach misleads video-based large language models into generating inaccurate captions. Besides, Mutes prompt these models to produce shorter and even NULL captions, lacking descriptive detail. The answers in the figure are excerpts.}
    \label{fig:intro}
\end{figure}
In this paper, we aim to protect personal videos from being unauthorizedly and automatedly annotated by video-based LLMs. To tackle this significant and challenging issue, we propose two series of protective and imperceptible watermarks added to videos. In each scenario, we propose two methods in order to add diversity and adaptability for widespread applications.

Specifically, we aim to mislead video-based LLMs into generating entirely incorrect annotations in the first scenario. To achieve this, we propose two methods, Rambling-F and Rambling-L, from different perspectives. Rambling-F operates at the feature level, increasing the discrepancy of video and LLM features. Rambling-L manipulates the logit-level by increasing the auto-regressive loss. These adversarial strategies serve to protect personal video content from unauthorized exploitation. Furthermore, fine-tuning text-to-video models on these adversarial video-text pairs will degrade the quality of generated videos.

The second scenario is designed to make the video-based LLMs effectively ``mute'', which means they are induced to generate shorter output on the given video. We also propose two methods, called Mute-S and Mute-N, relying on the probability of the End-of-Sequence (EOS) token, which, when generated, halts the LLM's inference process. Mute-S aims to cause the video-based LLMs to generate \textit{short} and incomplete text sequences. For example, the generated text is ``The video is a''. Mute-N aims to return a \textit{NULL} response by increasing the EOS probability of the first generated token. 
Overall, these methods can either fragment the sentences or result in a null output, preventing meaningful information from being leveraged by the video-based LLMs. These disruptions in the video-text pairs can then influence the text-to-video model's training and reduce your information leakage.

Our two scenarios are evaluated across three datasets and three different video-based LLMs, providing comprehensive validation of our methods' effectiveness and adaptability. The text output demonstrates that our Ramblings method can effectively induce the video-based LLMs to generate completely incorrect annotations for the videos, while our Mutes method successfully limits the model's output, reducing the amount of useful information generated. Additionally, we use the annotated video-text pairs to fine-tune a text-to-video model, and the results clearly show that the model's performance degrades by our watermarking methods. Finally, a series of ablation studies, including prompt transferability, further validate the effectiveness and robustness of our proposed methods.

In summary, our contribution can be outlined as follows:
\begin{itemize}
    \item We investigate the security and privacy risks faced by video annotation with video-based LLMs. To address these risks, we introduce two series of novel video watermarks that prevent unauthorized automated annotation. And we emphasize that the video-text pairs' quality influenced by our watermarks is important for downstream tasks such as text-to-video models.
    
    \item Our two series of watermarking methodologies demonstrate distinct strategies for disrupting video-based LLMs: inducing either inaccurate or short captions. These methodologies expand the scope of video privacy protection techniques and establish a foundation for developing robust defenses against unauthorized content analysis.
    
    \item Through extensive experiments, our methods have been proven to perform effectively across three state-of-the-art video-based LLMs, namely Video-ChatGPT~\cite{maaz2023video}, Video-LLaMA~\cite{zhang2023video}, and Video-Vicuna (change the backbone of Video-LLaMA), as well as three diverse datasets: OpenVid-1M~\cite{nan2024openvid}, MSR-VTT~\cite{xu2016msr}, and WebVid-10M~\cite{bain2021frozen}. These demonstrate the robustness and generalization capability of our approach, showing its ability to adapt to various models and datasets while consistently preventing unauthorized video annotation.
\end{itemize}

\section{Related Work}
\label{sec:related}

\subsection{Video-based Large Language Models}
The recent advancements in video-based LLMs~\cite{zhang2023video,huang2024vtimellm,maaz2023video,tan2024koala,xu2024pllava,yang2023vid2seq} have significantly enhanced their capabilities in various video-related tasks~\cite{zhang2023relational,zhang2024learning,wang2024efficient}. These models, such as Video-ChatGPT~\cite{maaz2023video} and Video-LLaMA~\cite{zhang2023video}, effectively leverage both spatial and temporal features from video data and the generation ability of LLMs, enabling them to achieve impressive zero-shot performance. The integration of the video modality into  LLMs allows for visual context-aware interaction, surpassing the capabilities of existing LLMs. The impressive performance of these models has led to their widespread use in generating dense captions~\cite{xu2024pllava} for unannotated videos, which can be used subsequently in downstream tasks, such as text-to-video generation models~\cite{chen2023videocrafter1,guo2023animatediff,lin2024animatediff,ma2024latte,wang2024animatelcm}. However, the widespread use of video-based LLMs raises privacy concerns regarding the potential for the unauthorized annotation of personal videos. As these models become more prevalent, it is crucial to address the ethical implications involving sensitive or private video content. To address these issues, we propose protective video watermarking methods that can prevent personal video from unauthorized and automated annotation by video-based LLMs.

\subsection{Adversarial Attack}

Adversarial attacks~\cite{chen2022nicgslowdown,bai2020targeted,he2023generating,gao2024adversarial,zong2023fool,dong2023robust,wu2023defenses,wang2024stop,fang2024clip,fang2024one,gao2024inducing,luo2024image,li2024video} typically craft imperceptible adversarial perturbations that can mislead the models into making incorrect predictions. With the development of multi-modal LLMs, researchers have begun exploring how these attacks can be applied in more complex generation settings. Concretely, recent studies show that adversarial attacks can be used to jailbreak multi-modal LLMs, causing them to generate harmful content~\cite{qi2023visual} or induce targeted captions~\cite{zhao2024evaluating}.
In contrast to these adversarial attacks, which aim to manipulate the model to an attacking goal, we adopt them as a pioneering watermarking strategy. It is designed to protect the privacy and security of video content within these models. Specifically, our watermarking approach is crafted to safeguard the video data against unauthorized annotation.

\section{Methodology}
\label{sec:method}

\begin{figure*}
  \centering\includegraphics[trim={30pt 550pt 30pt 0pt},clip,width=1\linewidth]{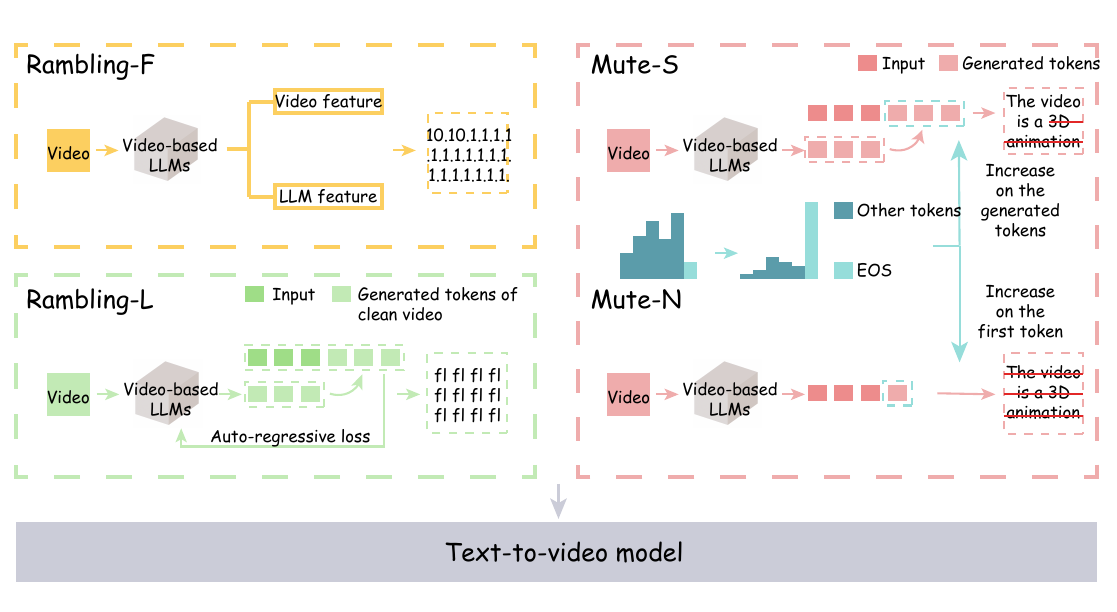}
  \vspace{-2em}
    \caption{The pipeline of Ramblings and Mutes. While Rambling-F focuses on feature-level perturbations of the original content, Rambling-L increases auto-regressive loss to manipulate video-based LLMs into generating incorrect descriptions. Mutes, on the other hand, tend to increase the probability of the EOS token. Finally, these video-text pairs can influence downstream tasks, such as the capacity of text-to-video models fine-tuned on these data.}
    \label{fig:v1}
\end{figure*}

\subsection{Overview}
To prevent unauthorized annotation by video-based LLMs, we propose two protective scenarios.
First, video-based LLMs can be misled to generate inaccurate captions that significantly diverge the video content when provided with protected video input. Second, video-based LLMs are induced to generate shorter, less detailed captions compared to those generated from clean videos which means less information leakage. Fine-tuning text-to-video models on such video-text pairs can degrade the quality of generated videos, due to the low quality of video-text pairs. For each scenario, we propose to adopt imperceptible adversarial perturbations as watermarks to protect videos from unauthorized annotation.

\subsection{Threat Model}

\textbf{Goals and Capabilities.} The goal is to craft an imperceptible adversarial perturbation as a protective watermark for the video, which can induce video-based LLMs to generate totally incorrect or shorter sequences during the victim model’s deployment. Following the most commonly used constraint for the involved perturbation~\cite{carlini2019evaluating}, it is restricted within a predefined magnitude in the $l_p$ norm, ensuring perturbation is difficult to detect.

\noindent \textbf{Knowledge and Background.} Following previous works~\cite{qi2024visual,gao2024inducing}, we assume that the victim video-based LLMs can be accessed in full knowledge, including architectures and parameters. Additionally, we consider a more challenging scenario where the victim video-based LLMs are inaccessible, as detailed in the Appendix.

\subsection{Problem Formulation}
To safeguard the video from unauthorized annotation by video-based LLMs, we propose to craft adversarial perturbations that can be added to a sequence of frames from the original videos. Given an original video $\bm{x}$ and an adversarial video $\bm{x}'$ with a watermark, the optimization problem can be formulated as follows:
\begin{equation}
\begin{aligned}
    &\arg \max_{\bm{x}'}\mathcal{L}(\bm{x}'), \\
    s.t&. \left || \bm{x}'-\bm{x} \right ||_\infty < \epsilon,
\end{aligned}
\end{equation}
where $\mathcal{L}(\cdot)$ denotes the loss objective function to achieve different protective scenarios. To ensure the stealthiness of the adversarial perturbations, we introduce the $l_\infty$ restriction within a value of $\epsilon$. For optimization, the projected gradient descent (PGD) algorithm~\cite{madry2017towards} is adopted to update the adversarial perturbations at each iteration.

Typically, when a video $\bm{x}$ and an input prompt $c_\text{in}$ are provided as inputs to the video-based LLMs $\mathcal{F}(\cdot)$, the video-based LLMs first utilize vision encoders $g(\cdot)$ to extract visual features from the video. These extracted video features are then fed into the base LLMs $h(\cdot)$ with $c_\text{in}$ as inputs, which generate the hidden states from its final layer. Finally, the hidden states are projected into the probability distribution of all generated tokens, which can obtain a sequence of generated output tokens. We denote $f_i(\cdot)$ as the probability distribution after the $\operatorname{Softmax(\cdot)}$ layer over the $i$-th generated token. 

\subsection{Ramblings for Incorrect Generation of Video-based LLMs}
\label{ramblings}
In this section, we optimize adversarial videos that can mislead video-based LLMs to generate incorrect captions. To achieve this goal, we develop two different methods for video watermarks from the feature level and the logit level of the video-based LLMs, dubbed Rambling-F and Rambling-L.

\noindent \textbf{Rambling-F (Feature).} To induce the video-based LLMs to generate incorrect captions, we propose a feature-level adversarial strategy called Rambling-F. The core idea behind this approach is to deviate the features of adversarial videos from the original ones in the latent representation space. During the inference stage, the video-based LLMs extract visual features from the input video and also generate LLM features from the combination of the input video and text queries. Hence, in the Rambling-F, we propose to increase the $l_2$ distance for both the video features and the LLM features between the original and adversarial videos. As a result, the disrupted features will result in the model's inability to accurately generate the captions correlated with the videos. The loss objective function for the Rambling-F can be formulated as follows:
\begin{equation}
\begin{aligned}    
    \mathcal{L}_{video}(\bm{x}'&)=\frac{1}{n}\displaystyle\sum_{i=1}^{n}(g(\bm{x'})_{i}-g(\bm{x})_{i})^2,\\
    \mathcal{L}_{LLM}(\bm{x}')=&\frac{1}{m}\displaystyle\sum_{j=1}^{m}(h(\bm{x'}, c_\text{in})_{j}-h(\bm{x}, c_\text{in})_{j})^2,\\
    \mathcal{L}_{RF}(\bm{x}')=&\alpha*\mathcal{L}_{video}(\bm{x}')+\beta*\mathcal{L}_{LLM}(\bm{x}'),
\end{aligned}
\label{eq:ramblings-F}
\end{equation}
where $n$ and $m$ are the total number of elements in the video and LLM features, respectively. In addition, $g(\cdot)_{i}$ represents the $i$-th dimension of video features and $h(\cdot)_{j}$ represents the $j$-th dimension of LLM features. 

\noindent \textbf{Rambling-L (Logit).} 
In addition to the feature-level Rambling-F attack, we propose another adversarial strategy called Rambling-L, which manipulates the logit-level outputs, \textit{i.e.}, probability distributions over the generated tokens. Typically, based on the probability distribution over generated tokens, video-based LLMs can generate one token at one time in an auto-regressive manner to obtain the caption. Therefore, for the generation of incorrect captions, we can increase the distance between the adversarial video $\bm{x}'$ and the original ground-truth caption $y$ generated from the original video $\bm{x}$ by the auto-regressive loss $\mathcal{L}_{ar}(\cdot)$. The loss objective function for the Rambling-L attack can be expressed as follows:
\begin{equation}
    \mathcal{L}_{RL}(\bm{x}')=\mathcal{L}_{ar}(\mathcal{F}(\bm{x}', c_\text{in}),y).
\label{eq:ramblings-L}
\end{equation}
By increasing \cref{eq:ramblings-L}, we can shift the model's token probability distribution away from the correct captions, causing the video-based LLMs to generate incorrect textual outputs.

\subsection{Mutes for Shorter even NULL Generation of Video-based LLMs}
\label{mute} In this section, we craft adversarial videos that can induce video-based LLMs to generate shorter captions or even no captions at all, including Mute-S and Mute-N. To this end, we propose manipulating the probability of the End-of-Sequence (EOS) token of video-based LLMs for two protective goals.

\noindent \textbf{Mute-S (Short Caption).} 
During the auto-regressive generation process of video-based LLMs, the occurrence of an End-of-Sequence (EOS) token serves as a signal to stop generating further tokens. 
To decrease the length of generated sequences, one straightforward approach is to induce the occurrence of the EOS token during the prediction process. However, the auto-regressive prediction is a non-deterministic random process~\cite{gao2024inducing}, making it challenging to directly determine the exact position where the EOS token is generated. To address this challenge, we propose Mute-S, which aims to increase the probability of the EOS token at all positions of generated sequences $y_\text{out}$. The loss function can be formulated as follows:
\begin{equation}
\begin{aligned}
\mathcal{L}_{MS}(\bm{x}')=\frac{1}{N} \sum_{i=1}^N f_{i}^{\mathrm{EOS}}\left(\bm{x}', c_\text{in} \oplus y_\text{out} \right),
\end{aligned}
\label{eq:Mute-S}
\end{equation}
where $f_{i}^{\mathrm{EOS}}(\cdot)$ is EOS token probability of the probability distribution after the $\operatorname{Softmax}(\cdot)$ layer over the $i$-th generated token and $N$ is the length of generated sequences $y_\text{out}$. Note that $y_\text{out}$ is updated at each iteration during the optimization. By increasing \cref{eq:Mute-S}, Mute-S can effectively manipulate video-based LLMs to terminate the caption generation process in advance, resulting in shorter and potentially less informative captions.

\noindent \textbf{Mute-N (NULL Caption).} 
Considering a more extreme case, we can directly induce video-based LLMs to become highly confident in predicting the EOS token as the very first token during generation, resulting in generating NULL or empty captions. To this end, we propose Mute-N, which increases the probability of the EOS token at the first predicted token position, rather than optimizing the mean probability of the EOS token across generated sequences as Mute-S. 
Given the auto-regressive generation process of video-based LLMs, we can force video-based LLMs to approach the EOS token in an auto-regressive manner. The loss function for Mute-N can be formulated as follows: 
\begin{equation}
    \mathcal{L}_{MN}(\bm{x}')=-\mathcal{L}_{ar}(\mathcal{F}(\bm{x}', c_\text{in}),\texttt{[EOS]}),
\label{eq:Mute-N}
\end{equation}
where $\mathcal{L}_{ar}(\cdot)$ denotes the 
auto-regressive loss function and \texttt{[EOS]} is the EOS token. By optimizing \cref{eq:Mute-N}, we can fool the video-based LLMs into generating the EOS token at the beginning, causing the generation process of video-based LLMs to halt immediately.

\begin{table*}
\begin{minipage}{\textwidth}
\caption{The CLIP score and BLEU between texts annotated by three video-based LLMs: Video-ChatGPT, Video-LLaMA, and Video-Vicuna, and clean captions in datasets are measured on three datasets: OpenVid-1M, MSR-VTT, and WebVid-10M. The best results are highlighted in \textbf{bold}.}
\label{tab:main result1}
\vspace{-1em}
\centering
\setlength\tabcolsep{10pt}{
\begin{tabular}{@{}ll|cc|cc|cc@{}}
\toprule
\multirow{3}{*}{Base model} & \multicolumn{1}{l|}{\multirow{3}{*}{Method}} & \multicolumn{2}{c|}{OpenVid-1M} & \multicolumn{2}{c|}{MSR-VTT} & \multicolumn{2}{c}{WebVid-10M}\\ 

 & & CLIP Score & BLEU & CLIP Score & BLEU & CLIP Score & BLEU   \\
\midrule 
\multirow{5}{*}{Video-ChatGPT} & Original & 0.76221 & 0.01703 & 0.67383 & 0.02405 & 0.60938 & 0.01040\\
& Noise  & 0.75781 & 0.01398 & 0.67529 & 0.02671 & 0.61377 & 0.01123\\
& Rambling-F & 0.66797 & 0.01197 & 0.60352 & 0.01561 & 0.49561 & 0.00514\\
& Rambling-L & \textbf{0.62744} & \textbf{0.00271} & \textbf{0.60303} & \textbf{0.00350} & \textbf{0.48315} & \textbf{0.00169}\\
\midrule
\multirow{5}{*}{Video-LLaMA} & Original & 0.78760 & 0.05121 & 0.62354 & 0.00482 & 0.60889 & 0.00316\\
& Noise & 0.79980 & 0.04761 & 0.62256 & 0.00472 & 0.60156 & 0.00292\\
& Rambling-F & \textbf{0.58301} & \textbf{0.01600} & 0.49292 & 0.00350 & \textbf{0.42896} & 0.00205\\
& Rambling-L & 0.60938 & 0.02068 & \textbf{0.49219} & \textbf{0.00313} & 0.43384 & \textbf{0.00189}\\
\midrule
\multirow{5}{*}{Video-Vicuna} & Original & 0.75732 & 0.01969 &0.62158 & 0.02604 & 0.60107 & 0.01332\\
& Noise & 0.76367 & 0.02275 &0.61768 & 0.02211 & 0.60107 & 0.01435\\
& Rambling-F & \textbf{0.62500}& 0.02317 & 0.50732 & \textbf{0.00784} & \textbf{0.42310} & \textbf{0.00363}\\
& Rambling-L & 0.63525 & \textbf{0.01644} & \textbf{0.50195} & 0.00908 & 0.44189 & 0.00449\\
\bottomrule
\end{tabular}}
\end{minipage}
\end{table*}

\begin{table*}
\begin{minipage}{\textwidth}
\caption{The length and EOS rate of texts annotated by three video-based LLMs: Video-ChatGPT, Video-LLaMA, and Video-Vicuna are measured on three datasets: OpenVid-1M, MSR-VTT, and WebVid-10M. The best results are highlighted in \textbf{bold}.}
\label{tab:main result2}
\vspace{-1em}
\centering
\setlength\tabcolsep{10.5pt}{
\begin{tabular}{@{}ll|cc|cc|cc@{}}
\toprule
\multirow{2}{*}{Base model} & \multicolumn{1}{l|}{\multirow{2}{*}{Method}} & \multicolumn{2}{c|}{OpenVid-1M} & \multicolumn{2}{c|}{MSR-VTT} & \multicolumn{2}{c}{WebVid-10M}\\ 
 & & Length & EOS Rate(\%) & Length & EOS Rate(\%) & Length & EOS Rate(\%)\\
\midrule 
\multirow{5}{*}{Video-ChatGPT} & Original & 30.47 & 0.0 & 30.29 & 0.0 & 31.21 & 0.0\\
& Noise & 30.20 & 0.0 & 31.24 & 0.0 & 24.28 & 0.0\\
& Mute-S & 13.60 & 11.0 & 12.28 & 13.0 & 11.39 & 21.0\\
& Mute-N & \textbf{0.00} & \textbf{100.0} & \textbf{0.00} & \textbf{100.0} & \textbf{0.00} & \textbf{100.0} \\
\midrule
\multirow{5}{*}{Video-LLaMA} & Original & 203.54 & 0.0 & 208.62 & 0.0 & 214.49 & 0.0\\
& Noise &  224.23 & 0.0 & 213.47 & 0.0 & 209.30 & 0.0\\
& Mute-S &  11.58 & 7.0 & 17.28 & 13.0 & 37.74 & 9.0 \\
& Mute-N &  \textbf{0.00} & \textbf{100.0} & \textbf{0.00} & \textbf{100.0} & \textbf{0.00} & \textbf{100.0}\\
\midrule
\multirow{5}{*}{Video-Vicuna} & Original & 32.90 & 0.0 & 46.54 & 0.0 & 28.64 & 0.0\\
& Noise & 38.27 & 0.0 & 40.08 & 0.0 & 34.37 & 0.0\\
& Mute-S & 7.38 & 1.0 & 6.45 & 0.0 & 5.56 & 0.0 \\
& Mute-N & \textbf{0.00} & \textbf{100.0} & \textbf{0.00} & \textbf{100.0}& \textbf{0.00} & \textbf{100.0}\\
\bottomrule
\end{tabular}}
\end{minipage}
\end{table*}
\section{Experiment}
\subsection{Implementation Details}
\textbf{Models and Datasets.} We elaborately select three state-of-the-art video-based LLMs as annotation models: Video-ChatGPT~\cite{maaz2023video}, Video-LLaMA~\cite{zhang2023video}, and Video-Vicuna (a variant of Video-LLaMA with a modified backbone). The default inference prompt template is used to obtain video annotations, with the user input prompt set to ``What is this video about?'' Detailed settings for these models are available in the Appendix. For datasets, we use data from OpenVid-1M~\cite{nan2024openvid}, MSR-VTT~\cite{xu2016msr}, and WebVid-10M~\cite{bain2021frozen}. They are all high-quality collections of videos paired with corresponding captions. By calculating similarity with the dataset captions, we can evaluate the performance of the annotations. To accommodate the limited memory of GPUs, each video is preprocessed to consist of 8 frames.

\noindent \textbf{Baselines and Setups.} 
We use original videos and videos added with random perturbations as baselines. For the setup, the perturbation magnitude is set to 16/255 using the $l_{\infty}$ norm to make the perturbation imperceptible, and the optimization step size is set to 1/255, using the projected gradient descent (PGD) algorithm~\cite{madry2017towards}. The optimization iteration is set to 200 for Ramblings and 500 for Mutes, respectively. For video-based LLM configurations, we set a sampling temperature of 0.2 and limit the output to a maximum of 512 tokens for three models and the others are default. Additional setup details are provided in the Appendix.

\noindent \textbf{Evaluation Metrics.} 
For the evaluation of Ramblings, we employ two metrics: CLIP score (RN101) and BLEU. CLIP score~\cite{radford2021learning} measures the similarity between the model’s post-attack inferences and the original dataset captions, with a higher score indicating greater similarity and better annotation performance. BLEU~\cite{papineni2002bleu} assesses the alignment between the model’s outputs and reference texts, where a higher BLEU indicates a closer match and improved annotation performance. Our adversarial objective is to reduce these scores.

For the evaluation of Mutes, we use two metrics: length and EOS rate (NULL caption rate). We compare the lengths of outputs generated from clean and adversarial videos to assess the effectiveness of the Mutes approach. A shorter length indicates less information leakage from the videos. The EOS rate measures how often video-based LLMs output the EOS token early, signifying that no information is leaked from the videos. A higher EOS rate means video-based LLMs output NULL captions more frequently.

\subsection{Main Results}
\label{main results}
As shown in \cref{tab:main result1}, the CLIP score and BLEU are similar between original videos and those with random noise across three models, indicating that random noise does not reduce annotation performance and is therefore ineffective at protecting videos from unauthorized annotation. In contrast, the evaluation scores for Ramblings in \cref{tab:main result1} show a substantial decrease compared to the baselines, indicating significant alterations in the semantic information generated by the video-based LLMs. These results demonstrate that adversarial attack on the feature level or logit level effectively misleads the models into generating incorrect annotations for the videos. And our two methods offer great adaptability and flexibility to accommodate various circumstances.

As shown in \cref{tab:main result2}, the videos with random noise are ineffective in protecting video content compared to the original videos, as the output length is high and the EOS rate is zero, indicating that a gradient optimization algorithm is necessary to achieve meaningful perturbation. In contrast,  our Mute-S effectively disrupts sentence integrity, for example, video-based LLMs may output fragmented text such as ``The video is a'' without forming a coherent description. Further examples are provided in the Appendix. Specifically, for Video-LLaMA, Mute-S reduces text length nearly tenfold, with the EOS rate around 10\%. Additionally, our Mute-S significantly reduces the length of generated texts while maintaining a relatively low EOS rate, representing an untargeted attack strategy. Meanwhile, our Mute-N successfully prompts the model to output NULL captions, meaning that video-based LLMs extract no information from the videos, achieving a targeted attack. This analysis underscores the exceptional effectiveness of both Mute-S and Mute-N, resulting in minimal information being extracted by video-based LLMs and ultimately protecting videos from unauthorized and automated annotation.

\begin{table*}
\begin{minipage}{\textwidth}
\caption{Prompt transferability: The CLIP score and BLEU between texts annotated by Video-LLaMA and clean captions are measured on OpenVid-1M dataset. The prompt used during the attack phase is ``What is this video about?''. After the attack, we apply three different prompts:``What is this video about?'', ``What happens in the video?'', and ``Can you describe the video in detail?'' to generate annotations for videos. The best results are highlighted in \textbf{bold}.}
\vspace{-1em}
\label{tab:prompt 1}
\centering
\setlength\tabcolsep{17.5pt}{
\begin{tabular}{@{}l|cc|cc|cc@{}}
\toprule
\multirow{3}{*}{Method} & \multicolumn{2}{c|}{\multirow{2}{*}{What is this video about?}} & \multicolumn{2}{c|}{\multirow{2}{*}{What happens in the video?}} & \multicolumn{2}{p{3cm}}{Can you describe the video in detail?}\\ 

 & CLIP Score & BLEU & CLIP Score & BLEU & CLIP Score & BLEU   \\
\midrule 
Original & 0.78760 & 0.05121 & 0.79346 & 0.04411 & 0.76758 & 0.03146 \\
Noise & 0.79980 & 0.04761 & 0.79688 & 0.04081 & 0.77197 & 0.03459\\
Rambling-F & \textbf{0.58301} & \textbf{0.01600} & \textbf{0.60010} & \textbf{0.01585} & \textbf{0.60645} & \textbf{0.01514}\\
Rambling-L  & 0.60938 & 0.02068 & 0.61279 & 0.01942 & 0.62500 & 0.01828\\
\bottomrule
\end{tabular}}
\end{minipage}
\end{table*}

\begin{table*}
\begin{minipage}{\textwidth}
\caption{Prompt transferability: The length and EOS rate of texts annotated by Video-LLaMA are measured on OpenVid-1M dataset. The prompt used during the attack phase is ``What is this video about?''. After the attack, we apply three different prompts: ``What is this video about?'', ``What happens in the video?'', and ``Can you describe the video in detail?'' to generate annotations for videos. The best results are highlighted in \textbf{bold}.}
\vspace{-1em}
\label{tab:prompt 2}
\centering
\setlength\tabcolsep{18pt}{
\begin{tabular}{@{}l|cc|cc|cc@{}}
\toprule
\multirow{3}{*}{Method} & \multicolumn{2}{c|}{\multirow{2}{*}{What is this video about?}} & \multicolumn{2}{c|}{\multirow{2}{*}{What happens in the video?}} & \multicolumn{2}{p{3cm}}{Can you describe the video in detail?}\\ 

 & Length & EOS Rate(\%) & Length & EOS Rate(\%) & Length & EOS Rate(\%)   \\
\midrule 
 Original & 203.54 & 0.0 & 221.31 & 0.0 & 299.12 & 0.0\\
Noise & 224.23 & 0.0 & 236.85 & 0.0 & 300.92 & 0.0\\
Mute-S  & 11.58 & 7.0 & \textbf{24.16} & 5.0 & \textbf{36.44} & 10.0\\
Mute-N & \textbf{0.00} & \textbf{100.0} & 64.04 & \textbf{73.0} & 230.74 & \textbf{19.0}\\
\bottomrule
\end{tabular}}
\end{minipage}
\end{table*}

\subsection{Influence on Downstream Work: Text-to-video Models}
To further demonstrate the impact of our watermarking methods on downstream tasks, we fine-tune text-to-video models using our annotated video-text pairs. After fine-tuning, we evaluate the model’s performance on established benchmarks to demonstrate that our protective watermarking methods can significantly degrade the model’s ability to generate high-quality videos. 

\begin{table}
  \centering
  \caption{The VQA\textsubscript{A} and VQA\textsubscript{T} are evaluated for AnimateDiff fine-tuned on different adversarial video-text pairs. All the video-text pairs are annotated by Video-LLaMA on the OpenVid-1M dataset. The best results are highlighted in \textbf{bold}.}
  \setlength\tabcolsep{15pt}
  \begin{tabular}{@{}lcc@{}}
    \toprule
    Method & VQA\textsubscript{A} & VQA\textsubscript{T} \\
    \midrule
    Original & 60.430 & 51.302 \\
    Noise & 46.718 & 35.392 \\
    Rambling-F & 39.383 & 28.231 \\
    Rambling-L & \textbf{38.042} & 27.394 \\
    Mute-S & 43.610 & \textbf{24.487} \\
    Mute-N & 43.468 & 31.009 \\
    \bottomrule
  \end{tabular}
  \label{tab:animate}
\end{table}

\noindent \textbf{Experiment Setups.} The text-to-video framework we used is AnimateDiff~\cite{guo2023animatediff}, which is based on diffusion model~\cite{rombach2022high}. The dataset we used to fine-tune the text-to-video model consists of video-text pairs annotated by Video-LLaMA based on OpenVid-1M. We follow the AnimateDiff training script. The pre-trained model used is stable-diffusion-v1-5~\footnote{https://huggingface.co/stable-diffusion-v1-5/stable-diffusion-v1-5}. For image fine-tuning in the AnimateDiff pipeline, the learning rate and training steps are set to $2 \times 10^{-6}$ and 500, respectively. During the training process of the motion module in the AnimateDiff pipeline, we set the learning rate to $5 \times 10^{-5}$ and the training steps to 10,000. For text-to-video generation, we set the guidance scale as 8 and the steps as 25. To evaluate conveniently, we still use stable-diffusion-v1-5 as the base model.

\noindent \textbf{Evaluation Metrics.} For the evaluation of the text-to-video models, we use two metrics: VQA\textsubscript{A} and VQA\textsubscript{T}. We adopt these metrics from EvalCrafter~\cite{liu2024evalcrafter} as benchmarks to evaluate the performance of text-to-video models fine-tuned on videos with annotations generated by video-based LLMs. Specifically, VQA\textsubscript{A} measures the aesthetic quality of generated videos, while VQA\textsubscript{T} assesses technical quality. Higher VQA\textsubscript{A} and VQA\textsubscript{T} indicate better quality of the generated videos by text-to-video models.

\noindent \textbf{Results.} As illustrated in \cref{tab:animate}, the VQA\textsubscript{A} and VQA\textsubscript{T}~\cite{liu2024evalcrafter} decrease when AnimateDiff is fine-tuned using adversarial video-text pairs compared to those with original or randomly perturbed video-text pairs. This notable reduction demonstrates that the quality of text-to-video models degrades significantly by our watermarking methods.

\subsection{Ablation Studies}
\label{ablation}
\textbf{Transferability.} Video-based LLMs generate corresponding responses according to different user prompts. In this section, we conduct the transferability experiments~\cite{luo2024image} to assess whether adversarial watermarks can protect videos from unauthorized annotation under varying user inference prompts. In this experiment, we adopt the inference prompts: ``What is this video about?'', ``What happens in the video?'', and ``Can you describe the video in detail?''

As shown in \cref{tab:prompt 1}, changing the inference prompt still results in a significant reduction in both CLIP score and BLEU when using our Ramblings method, indicating that Rambling-F and Rambling-L effectively protect videos from unauthorized annotation across varied prompts. Similarly, as shown in \cref{tab:prompt 2}, even with differing inference prompts, Mute-S and Mute-N significantly reduce the length of generated texts, preventing information leakage from the videos. Additionally, the EOS rate remains high, suggesting that Mute-N leads video-based LLMs to produce null captions. Notably, when the inference prompt differs from the prompt used during the attack phase, Mute-N shows a certain degree of performance reduction against Video-LLaMA. We speculate this may be due to Mute-N's focus on the EOS probability of a single token, potentially reducing its stability across different inference prompts. So, we introduce Mute-N2 in the Appendix to enhance its performance. These findings further emphasize the necessity of using both Mute-S and Mute-N, as Mute-S remains stable under different circumstances, while Mute-N performs optimally when the inference prompt matches the one used during the attack phase. Overall, the experiments above demonstrate that our methods have a certain degree of prompt transferability. 

\begin{figure}[t]
  \centering
  \begin{minipage}[b]{0.48\linewidth}
    \centering
    \includegraphics[trim={60pt 60pt 18pt 80pt},clip,width=\linewidth]{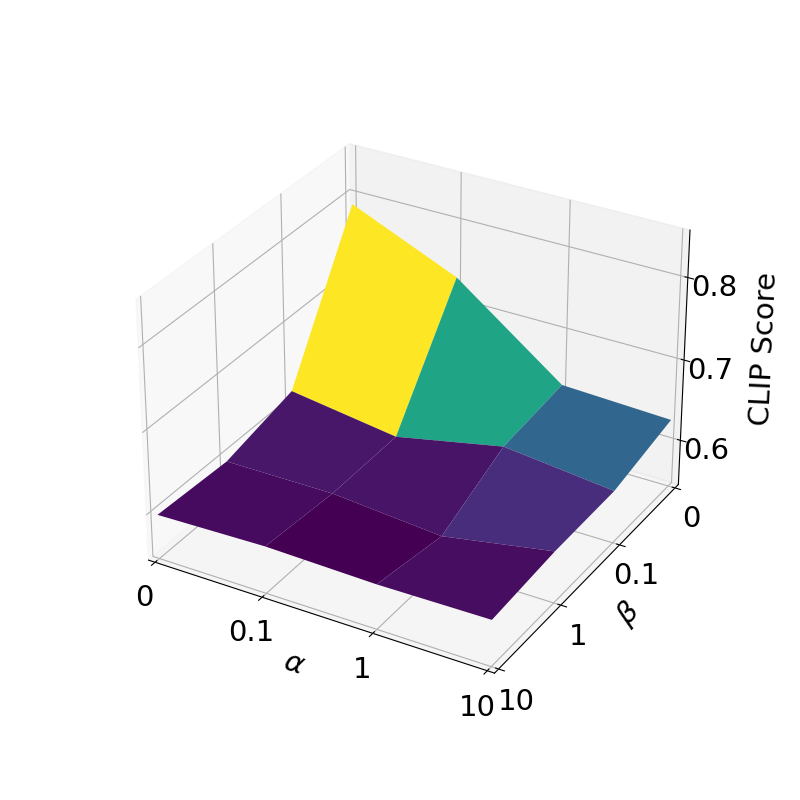}
  \end{minipage}
  \begin{minipage}[b]{0.48\linewidth}
    \centering
    \includegraphics[trim={60pt 60pt 18pt 80pt},clip,width=\linewidth]{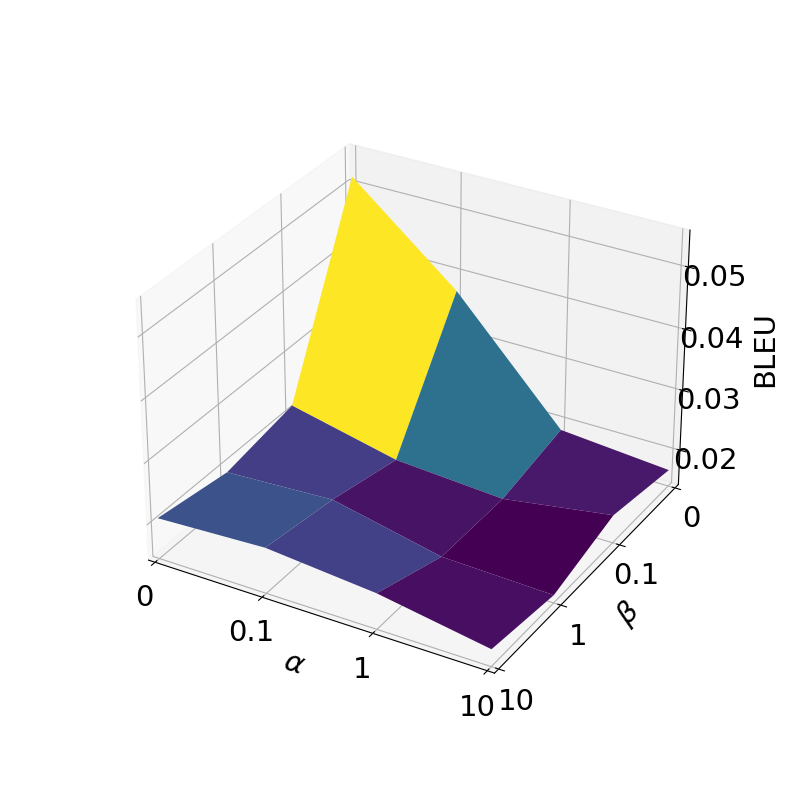}
  \end{minipage}
  \caption{Comparison of CLIP score (left) and BLEU (right) on Rambling-F to evaluate different parameters' influence. When the parameter is too small, the CLIP score and BLEU are high, indicating poor protective performance. Meanwhile, the evaluation scores decrease under $\alpha=1, \beta=1$ compared to $\alpha=1, \beta=0$ or $\alpha=0, \beta=1$, suggesting that the combination of feature losses is effective and necessary. All the videos are annotated by Video-LLaMA on the OpenVid-1M dataset. $\alpha=0,\beta=0$ represents original video.}
  \label{fig:para_sensi}
\end{figure}

\noindent \textbf{Ablation Study on Rambling-F.} Our Rambling-F integrates both $\mathcal{L}_{video}$ and $\mathcal{L}_{LLM}$. To thoroughly evaluate the individual and combined impact of each feature loss, we conduct a parameter experiment to explore the influence of $\alpha$ and $\beta$ in \cref{eq:ramblings-F}. As outlined in \cref{fig:para_sensi}, the combination of $\alpha=1,\beta=1$ outperforms the combination of $\alpha=1,\beta=0$ or $\alpha=0,\beta=1$ which means that it is effective to add these features together to get better protective performance. Additionally, larger values for these parameters generally lead to better performance in reducing annotation quality. However, when increased to a certain extent, the performance does not show improvement even reduction. These findings highlight the synergy between the two feature losses, suggesting that their integration enhances the method's ability to strengthen adversarial attacks and more effectively protect video content. And to get better performance, the proper parameters are needed. Actually, $\alpha=1,\beta=1$ is a suitable parameter combination. 

\begin{figure}[t]
  \centering
  \begin{minipage}[b]{0.45\linewidth}
    \centering
    \includegraphics[trim={20pt 0pt 22pt 20pt},clip,width=\linewidth]{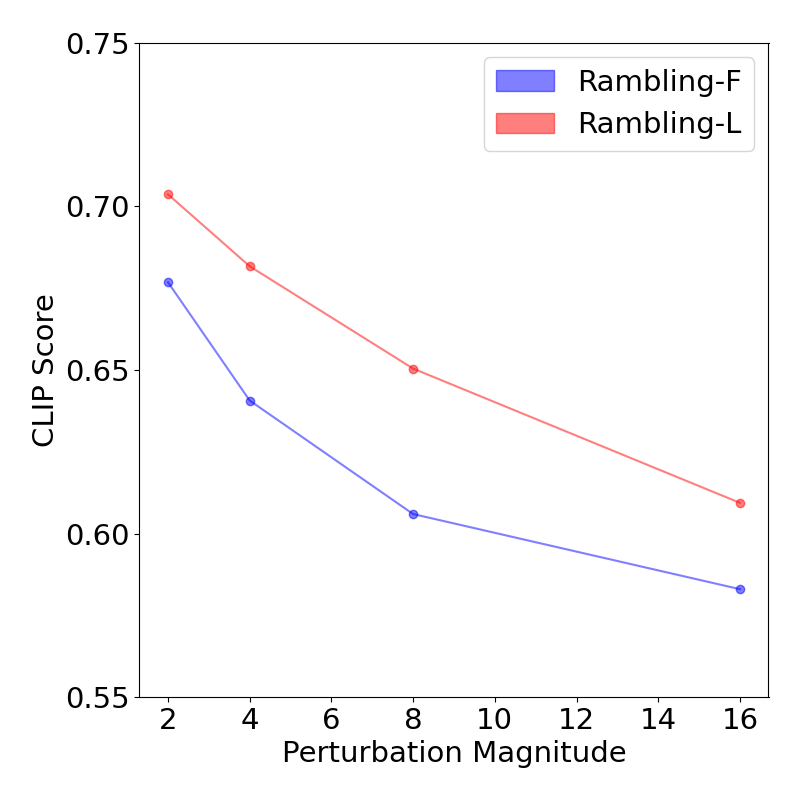}
    \caption{CLIP score about Ramblings on different perturbation magnitudes. When the perturbation magnitude is small, the CLIP score is higher, indicating poor protective performance for the videos. With the increase of perturbation magnitude, the protective performance of Ramblings is better.}
    \label{fig:pertur_clip}
  \end{minipage}
  \hfill
  \begin{minipage}[b]{0.48\linewidth}
    \centering
    \includegraphics[trim={0pt 0pt 0pt 20pt},clip,width=\linewidth]{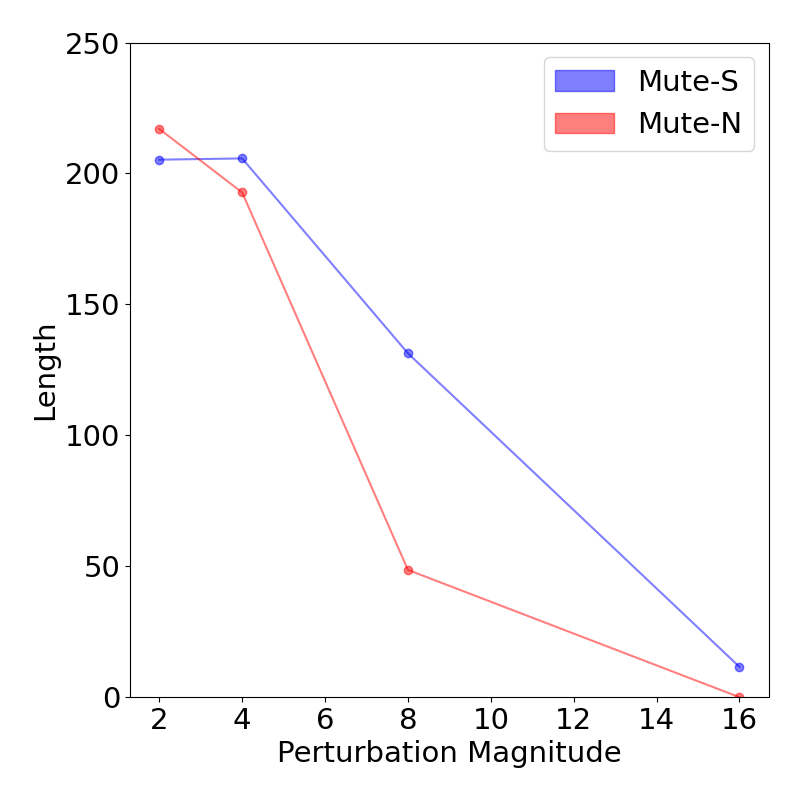}
    \caption{The length of the textual output for Mutes varies across different perturbation magnitudes. When the perturbation magnitude is small, the text length is longer, indicating more information leakage from the videos. With the increase of perturbation magnitude, the information leakage of Mutes is less.}
    \label{fig:pertur_len}
  \end{minipage}
  \hfill
\end{figure}

\noindent \textbf{Ablation Study on Perturbation Magnitude.}\label{perturbation} We conduct additional experiments with varying perturbation magnitudes to investigate their influence on adversarial performance. We select perturbation magnitudes of 2/255, 4/255, 8/255, and 16/255 for evaluation. As illustrated in \cref{fig:pertur_clip}, a perturbation magnitude of 2/255 yields the worst performance and a perturbation magnitude of 16/255 yields the best performance to protect the videos. Overall, an increase in the perturbation magnitude correlates with an enhancement in the Ramblings performance. Meanwhile, as illustrated in \cref{fig:pertur_len}, lower perturbation magnitude tends to result in increased output length. Overall, reduced perturbation magnitudes may compromise the performance of Ramblings and Mutes, yet they have the benefit of minimizing deviations from the original videos. Consequently, finding an optimal balance in perturbation magnitude is essential to achieve both effective performance in adversarial attacks and imperceptible deviations from original videos. More experiment results can be found in the Appendix. And all the videos used in experiments are annotated by Video-LLaMA on the OpenVid-1M dataset.

\section{Conclusion}
In this study, we have successfully developed two series of novel watermarking methods aimed at safeguarding video content from unauthorized and automated annotation by video-based large language models. Through our exploration of adversarial attacks, we propose two series of video watermarks named Ramblings and Mutes, respectively. On the one hand, Ramblings are designed to fool video-based LLMs into generating completely incorrect annotations. On the other hand, Mutes, which focus on the EOS token, can shorten annotations by video-based LLMs. Extensive experiments demonstrate the effectiveness of our watermarking methods in misleading or silencing video-based LLMs. Furthermore, evaluations of text-to-video models using different protective approaches show that our methods can impact downstream tasks. Finally, we hope that our Ramblings and Mutes can serve as a protective paradigm to help safeguard videos from unauthorized annotation. 

\section*{Ethics Statement}
Our proposed Ramblings and Mutes are designed solely for protective purposes not to be applied maliciously, and their application in this context is intended to benefit video data privacy. We hope that our work can inform both academic researchers and industry professionals on safeguarding personal video content from misuse.

\section*{Acknowledgement}
This work is supported in part by the National Natural Science Foundation of China, under Grant (62302309, 62171248), Shenzhen Science and Technology Program (JCYJ20220818101014030, JCYJ20220818101012025), and the PCNL KEY project (PCL2023AS6-1).

{
    \small
    \bibliographystyle{ieeenat_fullname}
    \bibliography{main}

\begin{thebibliography}{51}
\providecommand{\natexlab}[1]{#1}
\providecommand{\url}[1]{\texttt{#1}}
\expandafter\ifx\csname urlstyle\endcsname\relax
  \providecommand{\doi}[1]{doi: #1}\else
  \providecommand{\doi}{doi: \begingroup \urlstyle{rm}\Url}\fi

\bibitem[Bai et~al.(2020)Bai, Chen, Li, Wu, Guo, Xia, and Yang]{bai2020targeted}
Jiawang Bai, Bin Chen, Yiming Li, Dongxian Wu, Weiwei Guo, Shu-tao Xia, and En-hui Yang.
\newblock Targeted attack for deep hashing based retrieval.
\newblock In \emph{ECCV}, 2020.

\bibitem[Bai et~al.(2022)Bai, Chen, Gao, Wang, and Xia]{bai2022practical}
Jiawang Bai, Bin Chen, Kuofeng Gao, Xuan Wang, and Shu-Tao Xia.
\newblock Practical protection against video data leakage via universal adversarial head.
\newblock \emph{Pattern Recognition}, 131:\penalty0 108834, 2022.

\bibitem[Bain et~al.(2021)Bain, Nagrani, Varol, and Zisserman]{bain2021frozen}
Max Bain, Arsha Nagrani, G{\"u}l Varol, and Andrew Zisserman.
\newblock Frozen in time: A joint video and image encoder for end-to-end retrieval.
\newblock In \emph{ICCV}, 2021.

\bibitem[Brown et~al.(2020)Brown, Mann, Ryder, Subbiah, Kaplan, Dhariwal, Neelakantan, Shyam, Sastry, Askell, et~al.]{brown2020language}
Tom Brown, Benjamin Mann, Nick Ryder, Melanie Subbiah, Jared~D Kaplan, Prafulla Dhariwal, Arvind Neelakantan, Pranav Shyam, Girish Sastry, Amanda Askell, et~al.
\newblock Language models are few-shot learners.
\newblock In \emph{NeurIPS}, pages 1877--1901, 2020.

\bibitem[Carlini et~al.(2019)Carlini, Athalye, Papernot, Brendel, Rauber, Tsipras, Goodfellow, Madry, and Kurakin]{carlini2019evaluating}
Nicholas Carlini, Anish Athalye, Nicolas Papernot, Wieland Brendel, Jonas Rauber, Dimitris Tsipras, Ian Goodfellow, Aleksander Madry, and Alexey Kurakin.
\newblock On evaluating adversarial robustness.
\newblock \emph{arXiv preprint arXiv:1902.06705}, 2019.

\bibitem[Chen et~al.(2023)Chen, Xia, He, Zhang, Cun, Yang, Xing, Liu, Chen, Wang, et~al.]{chen2023videocrafter1}
Haoxin Chen, Menghan Xia, Yingqing He, Yong Zhang, Xiaodong Cun, Shaoshu Yang, Jinbo Xing, Yaofang Liu, Qifeng Chen, Xintao Wang, et~al.
\newblock Videocrafter1: Open diffusion models for high-quality video generation.
\newblock \emph{arXiv preprint arXiv:2310.19512}, 2023.

\bibitem[Chen et~al.(2022)Chen, Song, Haque, Liu, and Yang]{chen2022nicgslowdown}
Simin Chen, Zihe Song, Mirazul Haque, Cong Liu, and Wei Yang.
\newblock Nicgslowdown: Evaluating the efficiency robustness of neural image caption generation models.
\newblock In \emph{CVPR}, pages 15365--15374, 2022.

\bibitem[Deng et~al.(2024)Deng, Duan, Jin, Chang, Tian, Liu, Wang, Gao, Zou, Jin, et~al.]{deng2024deconstructing}
Chengyuan Deng, Yiqun Duan, Xin Jin, Heng Chang, Yijun Tian, Han Liu, Yichen Wang, Kuofeng Gao, Henry~Peng Zou, Yiqiao Jin, et~al.
\newblock Deconstructing the ethics of large language models from long-standing issues to new-emerging dilemmas: A survey.
\newblock \emph{arXiv preprint arXiv:2406.05392}, 2024.

\bibitem[Dong et~al.(2023)Dong, Chen, Chen, Fang, Yang, Zhang, Tian, Su, and Zhu]{dong2023robust}
Yinpeng Dong, Huanran Chen, Jiawei Chen, Zhengwei Fang, Xiao Yang, Yichi Zhang, Yu Tian, Hang Su, and Jun Zhu.
\newblock How robust is google's bard to adversarial image attacks?
\newblock \emph{arXiv preprint arXiv:2309.11751}, 2023.

\bibitem[Fang et~al.(2024{\natexlab{a}})Fang, Kong, Chen, Dai, Wu, and Xia]{fang2024clip}
Hao Fang, Jiawei Kong, Bin Chen, Tao Dai, Hao Wu, and Shu-Tao Xia.
\newblock Clip-guided generative networks for transferable targeted adversarial attacks.
\newblock In \emph{ECCV}, 2024{\natexlab{a}}.

\bibitem[Fang et~al.(2024{\natexlab{b}})Fang, Kong, Yu, Chen, Li, Xia, and Xu]{fang2024one}
Hao Fang, Jiawei Kong, Wenbo Yu, Bin Chen, Jiawei Li, Shutao Xia, and Ke Xu.
\newblock One perturbation is enough: On generating universal adversarial perturbations against vision-language pre-training models.
\newblock \emph{arXiv preprint arXiv:2406.05491}, 2024{\natexlab{b}}.

\bibitem[Gao et~al.(2024{\natexlab{a}})Gao, Bai, Bai, Yang, and Xia]{gao2024adversarial}
Kuofeng Gao, Yang Bai, Jiawang Bai, Yong Yang, and Shu-Tao Xia.
\newblock Adversarial robustness for visual grounding of multimodal large language models.
\newblock In \emph{ICLR Workshop on Reliable and Responsible Foundation Models}, 2024{\natexlab{a}}.

\bibitem[Gao et~al.(2024{\natexlab{b}})Gao, Bai, Gu, Xia, Torr, Li, and Liu]{gao2024inducing}
Kuofeng Gao, Yang Bai, Jindong Gu, Shu-Tao Xia, Philip Torr, Zhifeng Li, and Wei Liu.
\newblock Inducing high energy-latency of large vision-language models with verbose images.
\newblock In \emph{ICLR}, 2024{\natexlab{b}}.

\bibitem[Gao et~al.(2024{\natexlab{c}})Gao, Cai, Shuai, Gong, and Li]{gao2024embedding}
Kuofeng Gao, Huanqia Cai, Qingyao Shuai, Dihong Gong, and Zhifeng Li.
\newblock Embedding self-correction as an inherent ability in large language models for enhanced mathematical reasoning.
\newblock \emph{arXiv preprint arXiv:2410.10735}, 2024{\natexlab{c}}.

\bibitem[Gao et~al.(2024{\natexlab{d}})Gao, Pang, Du, Yang, Xia, and Lin]{gao2024denial}
Kuofeng Gao, Tianyu Pang, Chao Du, Yong Yang, Shu-Tao Xia, and Min Lin.
\newblock Denial-of-service poisoning attacks against large language models.
\newblock \emph{arXiv preprint arXiv:2410.10760}, 2024{\natexlab{d}}.

\bibitem[Guo et~al.(2024)Guo, Yang, Rao, Liang, Wang, Qiao, Agrawala, Lin, and Dai]{guo2023animatediff}
Yuwei Guo, Ceyuan Yang, Anyi Rao, Zhengyang Liang, Yaohui Wang, Yu Qiao, Maneesh Agrawala, Dahua Lin, and Bo Dai.
\newblock Animatediff: Animate your personalized text-to-image diffusion models without specific tuning.
\newblock In \emph{ICLR}, 2024.

\bibitem[He et~al.(2023)He, Liu, Li, Liang, Li, Jia, and Cao]{he2023generating}
Bangyan He, Jian Liu, Yiming Li, Siyuan Liang, Jingzhi Li, Xiaojun Jia, and Xiaochun Cao.
\newblock Generating transferable 3d adversarial point cloud via random perturbation factorization.
\newblock In \emph{AAAI}, 2023.

\bibitem[Huang et~al.(2024)Huang, Wang, Chen, Song, and Zhu]{huang2024vtimellm}
Bin Huang, Xin Wang, Hong Chen, Zihan Song, and Wenwu Zhu.
\newblock Vtimellm: Empower llm to grasp video moments.
\newblock In \emph{CVPR}, pages 14271--14280, 2024.

\bibitem[Li et~al.(2024)Li, Gao, Bai, Zhang, and Xia]{li2024video}
Jinmin Li, Kuofeng Gao, Yang Bai, Jingyun Zhang, and Shu-Tao Xia.
\newblock Video watermarking: Safeguarding your video from (unauthorized) annotations by video-based llms.
\newblock In \emph{ICML Workshop}, 2024.

\bibitem[Li et~al.(2022{\natexlab{a}})Li, Bai, Jiang, Yang, Xia, and Li]{li2022untargeted}
Yiming Li, Yang Bai, Yong Jiang, Yong Yang, Shu-Tao Xia, and Bo Li.
\newblock Untargeted backdoor watermark: Towards harmless and stealthy dataset copyright protection.
\newblock In \emph{NeurIPS}, 2022{\natexlab{a}}.

\bibitem[Li et~al.(2022{\natexlab{b}})Li, Zhu, Jia, Bai, Jiang, Xia, and Cao]{li2022move}
Yiming Li, Linghui Zhu, Xiaojun Jia, Yang Bai, Yong Jiang, Shu-Tao Xia, and Xiaochun Cao.
\newblock Move: Effective and harmless ownership verification via embedded external features.
\newblock \emph{arXiv preprint arXiv:2208.02820}, 2022{\natexlab{b}}.

\bibitem[Li et~al.(2023)Li, Zhu, Yang, Jiang, Wei, and Xia]{li2023black}
Yiming Li, Mingyan Zhu, Xue Yang, Yong Jiang, Tao Wei, and Shu-Tao Xia.
\newblock Black-box dataset ownership verification via backdoor watermarking.
\newblock \emph{IEEE Transactions on Information Forensics and Security}, 2023.

\bibitem[Lin and Yang(2024)]{lin2024animatediff}
Shanchuan Lin and Xiao Yang.
\newblock Animatediff-lightning: Cross-model diffusion distillation.
\newblock \emph{arXiv preprint arXiv:2403.12706}, 2024.

\bibitem[Liu et~al.(2024)Liu, Cun, Liu, Wang, Zhang, Chen, Liu, Zeng, Chan, and Shan]{liu2024evalcrafter}
Yaofang Liu, Xiaodong Cun, Xuebo Liu, Xintao Wang, Yong Zhang, Haoxin Chen, Yang Liu, Tieyong Zeng, Raymond Chan, and Ying Shan.
\newblock Evalcrafter: Benchmarking and evaluating large video generation models.
\newblock In \emph{CVPR}, pages 22139--22149, 2024.

\bibitem[Luo et~al.(2024)Luo, Gu, Liu, and Torr]{luo2024image}
Haochen Luo, Jindong Gu, Fengyuan Liu, and Philip Torr.
\newblock An image is worth 1000 lies: Transferability of adversarial images across prompts on vision-language models.
\newblock In \emph{ICLR}, 2024.

\bibitem[Ma et~al.(2024)Ma, Wang, Jia, Chen, Liu, Li, Chen, and Qiao]{ma2024latte}
Xin Ma, Yaohui Wang, Gengyun Jia, Xinyuan Chen, Ziwei Liu, Yuan-Fang Li, Cunjian Chen, and Yu Qiao.
\newblock Latte: Latent diffusion transformer for video generation.
\newblock \emph{arXiv preprint arXiv:2401.03048}, 2024.

\bibitem[Maaz et~al.(2024)Maaz, Rasheed, Khan, and Khan]{maaz2023video}
Muhammad Maaz, Hanoona Rasheed, Salman Khan, and Fahad~Shahbaz Khan.
\newblock Video-chatgpt: Towards detailed video understanding via large vision and language models.
\newblock In \emph{ACL}, 2024.

\bibitem[Madry et~al.(2017)Madry, Makelov, Schmidt, Tsipras, and Vladu]{madry2017towards}
Aleksander Madry, Aleksandar Makelov, Ludwig Schmidt, Dimitris Tsipras, and Adrian Vladu.
\newblock Towards deep learning models resistant to adversarial attacks.
\newblock \emph{arXiv preprint arXiv:1706.06083}, 2017.

\bibitem[Nan et~al.(2024)Nan, Xie, Zhou, Fan, Yang, Chen, Li, Yang, and Tai]{nan2024openvid}
Kepan Nan, Rui Xie, Penghao Zhou, Tiehan Fan, Zhenheng Yang, Zhijie Chen, Xiang Li, Jian Yang, and Ying Tai.
\newblock Openvid-1m: A large-scale high-quality dataset for text-to-video generation.
\newblock \emph{arXiv preprint arXiv:2407.02371}, 2024.

\bibitem[Papineni et~al.(2002)Papineni, Roukos, Ward, and Zhu]{papineni2002bleu}
Kishore Papineni, Salim Roukos, Todd Ward, and Wei-Jing Zhu.
\newblock Bleu: a method for automatic evaluation of machine translation.
\newblock In \emph{ACL}, pages 311--318, 2002.

\bibitem[Qi et~al.(2024{\natexlab{a}})Qi, Huang, Panda, Henderson, Wang, and Mittal]{qi2024visual}
Xiangyu Qi, Kaixuan Huang, Ashwinee Panda, Peter Henderson, Mengdi Wang, and Prateek Mittal.
\newblock Visual adversarial examples jailbreak aligned large language models.
\newblock In \emph{AAAI}, 2024{\natexlab{a}}.

\bibitem[Qi et~al.(2024{\natexlab{b}})Qi, Huang, Panda, Wang, and Mittal]{qi2023visual}
Xiangyu Qi, Kaixuan Huang, Ashwinee Panda, Mengdi Wang, and Prateek Mittal.
\newblock Visual adversarial examples jailbreak large language models.
\newblock In \emph{AAAI}, 2024{\natexlab{b}}.

\bibitem[Radford et~al.(2021)Radford, Kim, Hallacy, Ramesh, Goh, Agarwal, Sastry, Askell, Mishkin, Clark, et~al.]{radford2021learning}
Alec Radford, Jong~Wook Kim, Chris Hallacy, Aditya Ramesh, Gabriel Goh, Sandhini Agarwal, Girish Sastry, Amanda Askell, Pamela Mishkin, Jack Clark, et~al.
\newblock Learning transferable visual models from natural language supervision.
\newblock In \emph{ICML}, pages 8748--8763. PMLR, 2021.

\bibitem[Rombach et~al.(2022)Rombach, Blattmann, Lorenz, Esser, and Ommer]{rombach2022high}
Robin Rombach, Andreas Blattmann, Dominik Lorenz, Patrick Esser, and Bj{\"o}rn Ommer.
\newblock High-resolution image synthesis with latent diffusion models.
\newblock In \emph{CVPR}, pages 10684--10695, 2022.

\bibitem[Tan et~al.(2024)Tan, Sun, Hu, Wang, Deilamsalehy, Plummer, Russell, and Saenko]{tan2024koala}
Reuben Tan, Ximeng Sun, Ping Hu, Jui-hsien Wang, Hanieh Deilamsalehy, Bryan~A Plummer, Bryan Russell, and Kate Saenko.
\newblock Koala: Key frame-conditioned long video-llm.
\newblock In \emph{CVPR}, pages 13581--13591, 2024.

\bibitem[Touvron et~al.(2023)Touvron, Lavril, Izacard, Martinet, Lachaux, Lacroix, Rozi{\`e}re, Goyal, Hambro, Azhar, et~al.]{touvron2023llama}
Hugo Touvron, Thibaut Lavril, Gautier Izacard, Xavier Martinet, Marie-Anne Lachaux, Timoth{\'e}e Lacroix, Baptiste Rozi{\`e}re, Naman Goyal, Eric Hambro, Faisal Azhar, et~al.
\newblock Llama: Open and efficient foundation language models.
\newblock \emph{arXiv preprint arXiv:2302.13971}, 2023.

\bibitem[Wang et~al.(2024{\natexlab{a}})Wang, Huang, Shi, Bian, Song, Liu, and Li]{wang2024animatelcm}
Fu-Yun Wang, Zhaoyang Huang, Xiaoyu Shi, Weikang Bian, Guanglu Song, Yu Liu, and Hongsheng Li.
\newblock Animatelcm: Accelerating the animation of personalized diffusion models and adapters with decoupled consistency learning.
\newblock \emph{arXiv preprint arXiv:2402.00769}, 2024{\natexlab{a}}.

\bibitem[Wang et~al.(2025)Wang, Lian, Li, Wang, Feng, Chen, Zhang, and Xia]{wang2024efficient}
Jinpeng Wang, Niu Lian, Jun Li, Yuting Wang, Yan Feng, Bin Chen, Yongbing Zhang, and Shu-Tao Xia.
\newblock Efficient self-supervised video hashing with selective state spaces.
\newblock In \emph{AAAI}, 2025.

\bibitem[Wang et~al.(2024{\natexlab{b}})Wang, Han, Chen, Xue, Ding, Xiao, Tresp, Torr, and Gu]{wang2024stop}
Zefeng Wang, Zhen Han, Shuo Chen, Fan Xue, Zifeng Ding, Xun Xiao, Volker Tresp, Philip Torr, and Jindong Gu.
\newblock Stop reasoning! when multimodal llms with chain-of-thought reasoning meets adversarial images.
\newblock \emph{arXiv preprint arXiv:2402.14899}, 2024{\natexlab{b}}.

\bibitem[Wei et~al.(2024)Wei, Wang, Gao, Shao, Li, Wang, and Qin]{wei2024pointncbw}
Cheng Wei, Yang Wang, Kuofeng Gao, Shuo Shao, Yiming Li, Zhibo Wang, and Zhan Qin.
\newblock Pointncbw: Towards dataset ownership verification for point clouds via negative clean-label backdoor watermark.
\newblock \emph{IEEE Transactions on Information Forensics and Security}, 2024.

\bibitem[Wei et~al.(2022)Wei, Wang, Schuurmans, Bosma, Xia, Chi, Le, Zhou, et~al.]{wei2022chain}
Jason Wei, Xuezhi Wang, Dale Schuurmans, Maarten Bosma, Fei Xia, Ed Chi, Quoc~V Le, Denny Zhou, et~al.
\newblock Chain-of-thought prompting elicits reasoning in large language models.
\newblock In \emph{NeurIPS}, pages 24824--24837, 2022.

\bibitem[Wu et~al.(2023)Wu, Wei, Zhu, Zheng, Zhu, Zhang, Chen, Yuan, Liu, and Liu]{wu2023defenses}
Baoyuan Wu, Shaokui Wei, Mingli Zhu, Meixi Zheng, Zihao Zhu, Mingda Zhang, Hongrui Chen, Danni Yuan, Li Liu, and Qingshan Liu.
\newblock Defenses in adversarial machine learning: A survey.
\newblock \emph{arXiv preprint arXiv:2312.08890}, 2023.

\bibitem[Xu et~al.(2016)Xu, Mei, Yao, and Rui]{xu2016msr}
Jun Xu, Tao Mei, Ting Yao, and Yong Rui.
\newblock Msr-vtt: A large video description dataset for bridging video and language.
\newblock In \emph{CVPR}, 2016.

\bibitem[Xu et~al.(2024)Xu, Zhao, Zhou, Lin, Ng, and Feng]{xu2024pllava}
Lin Xu, Yilin Zhao, Daquan Zhou, Zhijie Lin, See~Kiong Ng, and Jiashi Feng.
\newblock Pllava: Parameter-free llava extension from images to videos for video dense captioning.
\newblock \emph{arXiv preprint arXiv:2404.16994}, 2024.

\bibitem[Yang et~al.(2023)Yang, Nagrani, Seo, Miech, Pont-Tuset, Laptev, Sivic, and Schmid]{yang2023vid2seq}
Antoine Yang, Arsha Nagrani, Paul~Hongsuck Seo, Antoine Miech, Jordi Pont-Tuset, Ivan Laptev, Josef Sivic, and Cordelia Schmid.
\newblock Vid2seq: Large-scale pretraining of a visual language model for dense video captioning.
\newblock In \emph{CVPR}, pages 10714--10726, 2023.

\bibitem[Zhang et~al.(2023{\natexlab{a}})Zhang, Li, and Bing]{zhang2023video}
Hang Zhang, Xin Li, and Lidong Bing.
\newblock Video-llama: An instruction-tuned audio-visual language model for video understanding.
\newblock \emph{arXiv preprint arXiv:2306.02858}, 2023{\natexlab{a}}.

\bibitem[Zhang et~al.(2018)Zhang, Isola, Efros, Shechtman, and Wang]{zhang2018perceptual}
Richard Zhang, Phillip Isola, Alexei~A Efros, Eli Shechtman, and Oliver Wang.
\newblock The unreasonable effectiveness of deep features as a perceptual metric.
\newblock In \emph{CVPR}, 2018.

\bibitem[Zhang et~al.(2023{\natexlab{b}})Zhang, Lu, Zhang, Nie, Yin, and Shen]{zhang2023relational}
Tong Zhang, Xiankai Lu, Hao Zhang, Xiushan Nie, Yilong Yin, and Jianbing Shen.
\newblock Relational network via cascade crf for video language grounding.
\newblock \emph{IEEE Transactions on Multimedia}, 26:\penalty0 8297--8311, 2023{\natexlab{b}}.

\bibitem[Zhang et~al.(2024)Zhang, Fang, Zhang, Gao, Lu, Nie, and Yin]{zhang2024learning}
Tong Zhang, Hao Fang, Hao Zhang, Jialin Gao, Xiankai Lu, Xiushan Nie, and Yilong Yin.
\newblock Learning feature semantic matching for spatio-temporal video grounding.
\newblock \emph{IEEE Transactions on Multimedia}, 2024.

\bibitem[Zhao et~al.(2023)Zhao, Pang, Du, Yang, Li, Cheung, and Lin]{zhao2024evaluating}
Yunqing Zhao, Tianyu Pang, Chao Du, Xiao Yang, Chongxuan Li, Ngai-Man~Man Cheung, and Min Lin.
\newblock On evaluating adversarial robustness of large vision-language models.
\newblock In \emph{NeurIPS}, 2023.

\bibitem[Zong et~al.(2024)Zong, Yu, Chavhan, Zhao, and Hospedales]{zong2023fool}
Yongshuo Zong, Tingyang Yu, Ruchika Chavhan, Bingchen Zhao, and Timothy Hospedales.
\newblock Fool your (vision and) language model with embarrassingly simple permutations.
\newblock In \emph{ICML}, pages 62892--62913, 2024.

\end{thebibliography}
}

\clearpage

\renewcommand\thesection{\Alph{section}}
\setcounter{section}{0}

\maketitlesupplementary

\appendix
\section{Overview}
The algorithm details are shown in \cref{app:algo}. Detailed implementation information, including model and method setups, can be found in \cref{app:imp}. In \cref{app:lenth distribution}, we analyze the length distribution of Mute-S to demonstrate its effectiveness in producing short sentences. \cref{app:perturbation} provides our ablation study on perturbation magnitude as a supplement for \cref{perturbation}. We conduct the prompt transferability experiment on non-annotation prompts to show our Mutes robustness in \cref{app:prompt}. Different inference parameters' influence is described in \cref{app:para}. Text perturbation's influence for fine-tuning text-to-video models when videos are clean is shown in \cref{app:text perturbation}. Then, we conduct a black-box attack experiment in \cref{app:black}. In \cref{app:explain}, we provide explanations of the different methods. We then perform adaptive evaluations in \cref{app:adaptive} to further demonstrate the robustness of our methods. Additionally, we optimize the prompt transferability of Mute-N in \cref{optimization of prompt} and evaluate the quality of adversarial videos in \cref{addition evaluation}. Finally, we present demos of different methods in \cref{app:visualization}. Additionally, for ease of processing video annotations, line breaks in the textual outputs are replaced with spaces in this paper. It is also worth noting that the perturbation shown in \cref{fig:intro} is illustrative and not the real perturbation, as the real perturbation contains negative values. The answers in \cref{fig:intro} and \cref{fig:v1} are also illustrative not the real answers. And the original video and clean video in this paper all mean the video without added perturbation.

\section{Algorithm Details}
\label{app:algo}
Our algorithms are detailed in \cref{eq:algorithm1,eq:algorithm2,eq:algorithm3,eq:algorithm4}. Here, $clip(\cdot)$ restricts the adversarial perturbation, while $G(\cdot)$ denotes the generation function of video-based LLMs.

We use auto-regressive loss in \cref{eq:Mute-N} for Mute-N. In video-based LLMs, auto-regressive loss typically relies on the cross-entropy loss function, which is composed of the log\_softmax and nll\_loss functions. When only the EOS token is used as the target, the effect of increasing the first generated token's EOS probability via the log\_softmax function is equivalent to the effect of \cref{eq:Mute-N}. For simplicity in program implementation, when programming, we directly manipulate the EOS probability of the first generated token using the log\_softmax function instead of computing the auto-regressive loss by passing label into the forward function.

\begin{algorithm}
\DontPrintSemicolon
\KwIn{original video $\bm{x}$, perturbation magnitude $\epsilon$, optimization step size $s$, input prompt $c_\text{in}$}
\KwOut{adversarial video $\bm{x'}$}
\BlankLine
Initialize perturbation: $\Delta$ 

\While{$i<T$}{
    $\bm{x'} \leftarrow \bm{x} + \Delta$\;
    
    $ \mathcal{L}_{video}(\bm{x'}),\mathcal{L}_{LLM}(\bm{x'}) \leftarrow \cref{eq:ramblings-F}$\
    
    $\mathcal{L} \leftarrow \alpha \times\mathcal{L}_{video}(\bm{x'}) + \beta \times \mathcal{L}_{LLM}(\bm{x'})$\

    $\Delta \leftarrow \Delta - s \times  sign(\nabla{\mathcal{L}})$\
      
    $clip(\bm{x},\Delta,\epsilon)$\

     $i \leftarrow i + 1$\
}
\Return{$\bm{x'}$}
\caption{Rambling-F}
\label{eq:algorithm1}
\end{algorithm}

\begin{algorithm}
\DontPrintSemicolon
\KwIn{original video $\bm{x}$, perturbation magnitude $\epsilon$, optimization step size $s$, input prompt $c_\text{in}$}
\KwOut{adversarial video $\bm{x'}$}
\BlankLine
Initialize perturbation: $\Delta$, clean caption: $y \leftarrow G(\bm{x},c_\text{in})$

\While{$i<T$}{
    $\bm{x'} \leftarrow \bm{x} + \Delta$\;
    
    $ \mathcal{L} \leftarrow \cref{eq:ramblings-L} $\
    
    $\Delta \leftarrow \Delta-s \times  sign(\nabla{\mathcal{L}})$\
      
    $clip(\bm{x},\Delta,\epsilon)$\

     $i \leftarrow i + 1$\
}
\Return{$\bm{x'}$}
\caption{Rambling-L}
\label{eq:algorithm2}
\end{algorithm}

\begin{algorithm}
\DontPrintSemicolon
\KwIn{original video $\bm{x}$, perturbation magnitude $\epsilon$, optimization step size $s$, input prompt $c_\text{in}$}
\KwOut{adversarial video $\bm{x'}$}
\BlankLine
Initialize perturbation: $\Delta$

\While{$i<T$}{
    $\bm{x'} \leftarrow \bm{x} + \Delta$\;
    
    $y_\text{out} \leftarrow G(\bm{x'},c_\text{in})$\
    
    $\mathcal{L} \leftarrow \cref{eq:Mute-S} $\

    $\Delta \leftarrow \Delta-s \times  sign(\nabla{\mathcal{L}})$\
      
    $clip(\bm{x},\Delta,\epsilon)$\

     $i \leftarrow i + 1$\
}
\Return{$\bm{x'}$}
\caption{Mute-S}
\label{eq:algorithm3}
\end{algorithm}

\begin{algorithm}
\DontPrintSemicolon
\KwIn{original video $\bm{x}$, perturbation magnitude $\epsilon$, optimization step size $s$, input prompt $c_\text{in}$}
\KwOut{adversarial video $\bm{x'}$}
\BlankLine
Initialize perturbation: $\Delta$

\While{$i<T$}{
    $\bm{x'} \leftarrow \bm{x} + \Delta$\;
        
    $\mathcal{L} \leftarrow \cref{eq:Mute-N} $\

    $\Delta \leftarrow \Delta-s \times  sign(\nabla{\mathcal{L}})$\
      
    $clip(\bm{x},\Delta,\epsilon)$\

     $i \leftarrow i + 1$\
}
\Return{$\bm{x'}$}
\caption{Mute-N}
\label{eq:algorithm4}
\end{algorithm}

\section{Implementation Details}
\label{app:imp}
Our adversarial experiments are run on NVIDIA GeForce RTX 3090, 24GB. We respectively use 100 videos from OpenVid-1M~\cite{nan2024openvid}, MSR-VTT~\cite{xu2016msr}, and WebVid-10M~\cite{bain2021frozen}. We preprocess the videos into 8 frames with 5 fps. And the video pictures in this paper are from these datasets. For AnimateDiff, we use 200 videos preprocessed into 8 frames with 5fps from OpenVid-1M.
\subsection{Model Setups}
\textbf{Video-ChatGPT.} We follow the offline demo settings from~\cite{maaz2023video} to configure Video-ChatGPT.

\noindent \textbf{Video-LLaMA.} We use Video-LLaMA-2-7B-Pretrained~\footnote{https://huggingface.co/DAMO-NLP-SG/Video-LLaMA-2-7B-Pretrained} to configure Video-LLaMA. We follow the demo for video without audio to configure.

\noindent \textbf{Video-Vicuna.} We change the backbone of Video-LLaMA from 
Llama-2-7b-chat~\footnote{https://huggingface.co/meta-llama} into Vicuna-7B~\footnote{https://huggingface.co/lmsys/vicuna-7b-delta-v0} and make it into Video-Vicuna. On the other hand, the vision-language branch we used is finetune-vicuna7b-v2~\footnote{https://huggingface.co/DAMO-NLP-SG/Video-LLaMA-Series}. We follow the demo for video without audio to configure.

\subsection{Method Setups}
\textbf{Rambling-F.} Three victim models share the same adversarial setups. The number of optimization iteration is set as 200 with a perturbation magnitude $\epsilon$ as 16/255. And the optimization step size is 1/255. Both the $\alpha$ and $\beta$ in the \cref{eq:ramblings-F} are set to 1.

\noindent \textbf{Rambling-L.} Three victim models share the same adversarial setups. The number of optimization iteration is set as 200 with a perturbation magnitude $\epsilon$ as 16/255. And the optimization step size is 1/255.

\noindent \textbf{Mute-S.} The number of optimization iteration is set as 500 with a perturbation magnitude $\epsilon$ as 16/255. We set the optimization step size as 1/255. To accelerate the optimization process, we set distinct iteration-breaking thresholds for the three victim models: a threshold of 3 for Video-ChatGPT and 4 for both Video-LLaMA and Video-Vicuna. If the loss falls below the threshold, the adversarial pipeline stops immediately. 

\noindent \textbf{Mute-N.} The number of optimization iteration is set as 500 with a perturbation magnitude $\epsilon$ as 16/255. We set the optimization step size as 1/255. To accelerate the optimization process, we set an iteration-breaking threshold of 0.3. If the loss falls below this threshold, the adversarial pipeline stops immediately.

\section{Additional Studies}
\label{app:abl}
\subsection{Length Distribution of Mute-S}\label{app:lenth distribution} To evaluate the detailed performance of Mute-S, which focuses on the EOS token, we analyze the distribution of text length generated by this method.

\begin{figure*}[htbp]
  \centering
  \begin{minipage}[b]{0.3\linewidth}
    \centering
    \includegraphics[width=\linewidth]{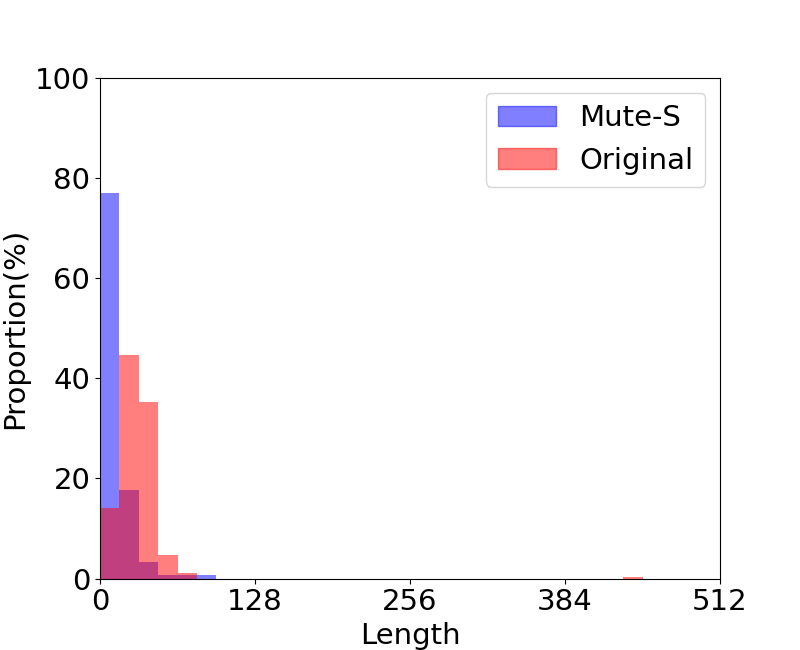}
    \caption{Length distribution of Mute-S on Video-ChatGPT.}
    \label{fig:gpt_length}
  \end{minipage}
  \hfill
  \begin{minipage}[b]{0.3\linewidth}
    \centering
    \includegraphics[width=\linewidth]{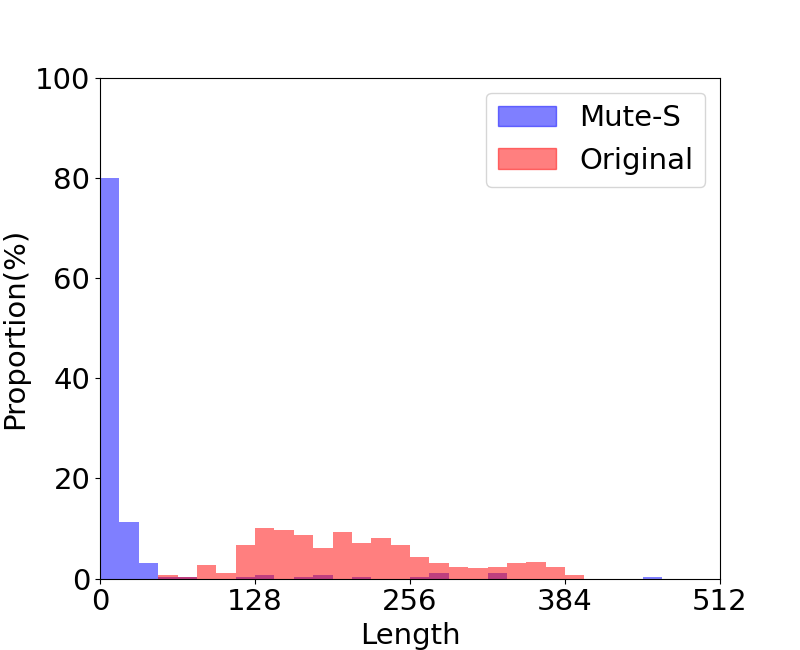}
    \caption{Length distribution of Mute-S on Video-LLaMA.}
    \label{fig:vllama_length}
  \end{minipage}
  \hfill
  \begin{minipage}[b]{0.3\linewidth}
    \centering
    \includegraphics[width=\linewidth]{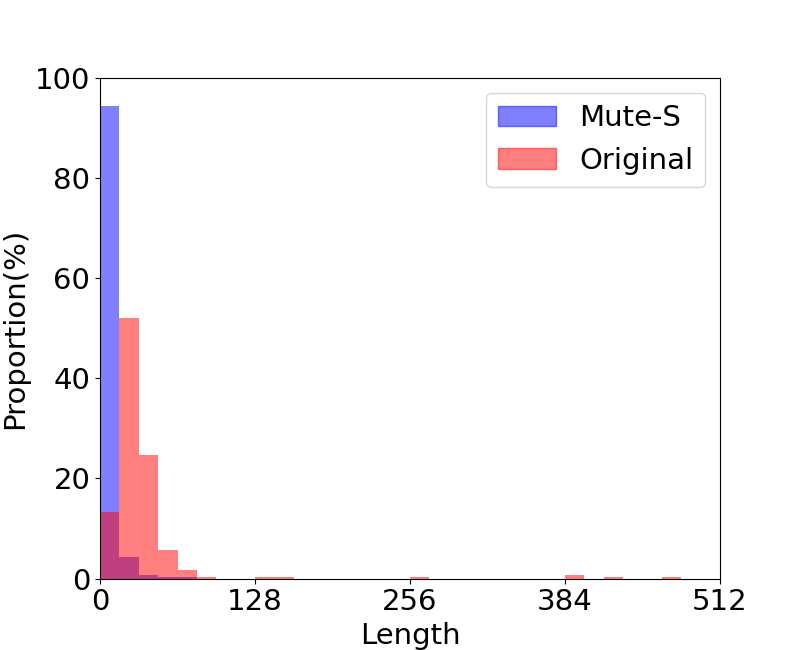}
    \caption{Length distribution of Mute-S on Video-Vicuna.}
    \label{fig:vicuna_length}
  \end{minipage}
\end{figure*}

\cref{fig:gpt_length}, \cref{fig:vllama_length}, and \cref{fig:vicuna_length} clearly illustrate that, with the addition of watermarks through Mute-S, the length distribution shifts noticeably to the left along the X-axis. This shift indicates a concentration of shorter text length, highlighting the effectiveness of Mute-S in reducing information leakage. Furthermore, we calculate the rate at which an alphabetic character appears in the final position of the output text, finding that 80.33\% of annotations generated by Video-LLaMA end in this manner. This suggests that sentences frequently lack a punctuation mark at the end, rendering them incomplete. Thus, Mute-S effectively disrupts text continuity, producing shorter and truncated outputs. In this section, the datasets we used are OpenVid-1M, MSR-VTT, and WebVid-10M.

\subsection{Ablation Study on Perturbation Magnitude}\label{app:perturbation} 
\begin{figure*}[htbp]
  \centering
  \begin{minipage}[b]{0.4\linewidth}
    \centering
    \includegraphics[trim={0 0 0 50},clip,width=\linewidth]{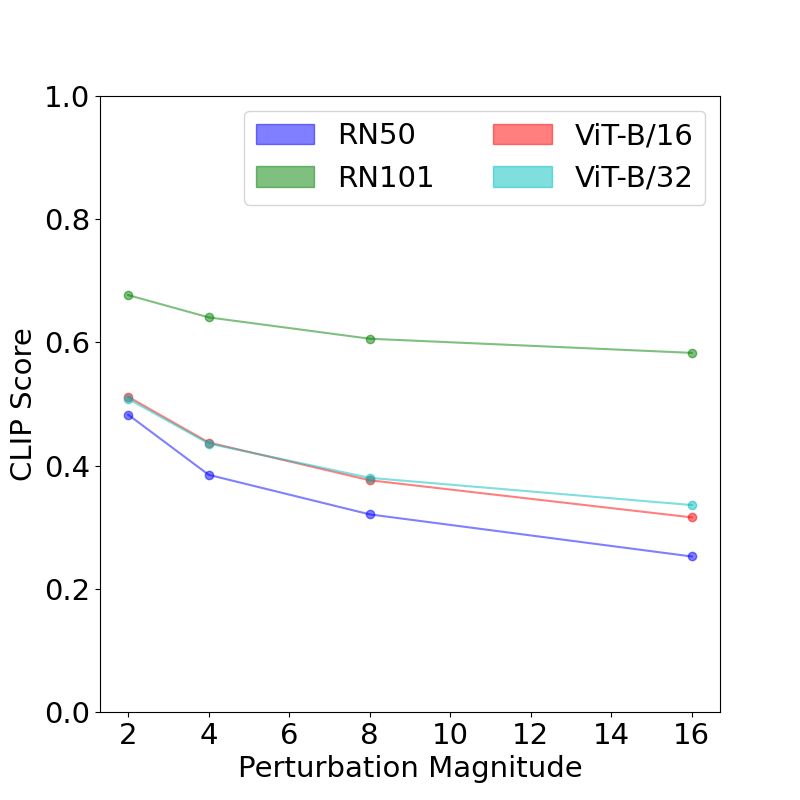}
    \caption{CLIP score about Rambling-F on different perturbation magnitudes.}
    \label{fig:perturb41}
  \end{minipage}
  \hfill
  \begin{minipage}[b]{0.4\linewidth}
    \centering
    \includegraphics[trim={0 0 0 50},clip,width=\linewidth]{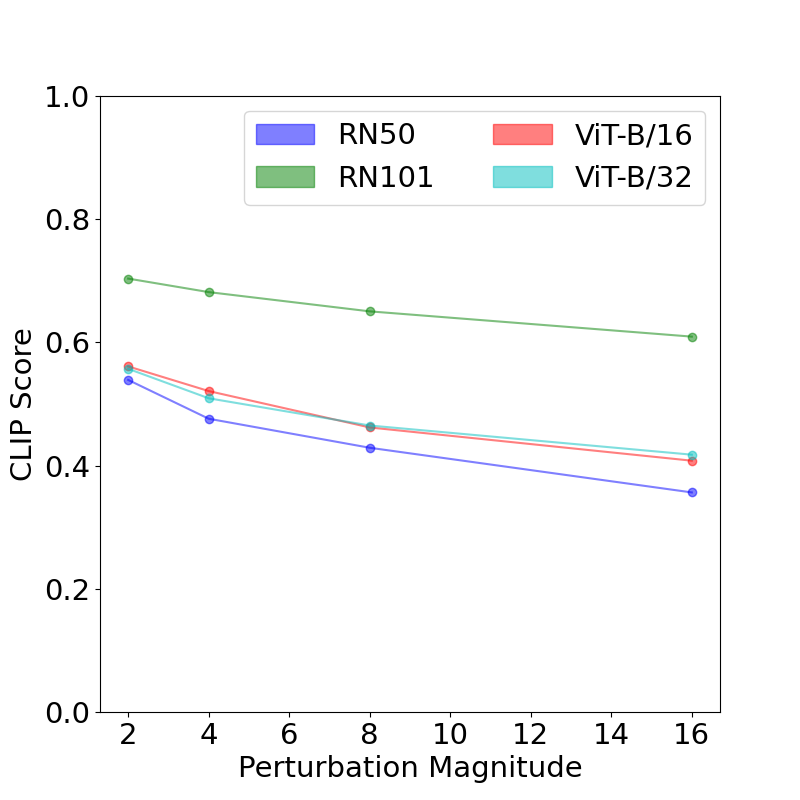}
    \caption{CLIP score about Rambling-L on different perturbation magnitudes.}
    \label{fig:perturb42}
  \end{minipage}
\end{figure*}

This section is a supplement for \cref{perturbation}. As shown in \cref{fig:perturb41}, \cref{fig:perturb42}, and \cref{tab:perturbation1}, a lower perturbation magnitude tends to result in a higher CLIP score and BLEU, indicating improved annotation performance and reduced protective effectiveness for the videos. Therefore, for Ramblings, it is crucial to select an appropriate perturbation magnitude to ensure its performance. \cref{tab:perturbation2} indicates that when the perturbation magnitude is too small, Mutes may become ineffective. We speculate that this is because Mutes rely on the EOS token's probability, and a slight increase may not be sufficient to prompt video-based LLMs to generate the EOS token. Therefore, it is important to avoid selecting a too-small perturbation magnitude for Mutes.

\begin{table*}
\begin{minipage}{\textwidth}
\caption{The CLIP score and BLEU between texts annotated by Video-LLaMA and clean captions are measured on OpenVid-1M dataset to explore the influence of different perturbation magnitudes $\epsilon$. The best results are highlighted in \textbf{bold}.}
\vspace{-1em}
\label{tab:perturbation1}
\centering
\setlength\tabcolsep{8pt}{
\begin{tabular}{@{}l|cc|cc|cc|cc@{}}
\toprule
\multirow{2}{*}{Method} & \multicolumn{2}{c|}{$\epsilon$ = 2} & \multicolumn{2}{c|}{$\epsilon$ = 4} & \multicolumn{2}{c|}{$\epsilon$ = 8} & \multicolumn{2}{c}{$\epsilon$ = 16}\\ 

 & CLIP Score & BLEU & CLIP Score & BLEU & CLIP Score & BLEU & CLIP Score & BLEU  \\
\midrule 
 Original & 0.78760 & 0.05121 & 0.78760 & 0.05121 & 0.78760 & 0.05121 & 0.78760 & 0.05121\\
Noise &0.79834 & 0.05041 & 0.79297 & 0.04500 & 0.79297 & 0.04595 & 0.79980 & 0.04761 \\
Rambling-F & \textbf{0.67676} & \textbf{0.02740} & \textbf{0.64062} &\textbf{0.02275} & \textbf{0.60596} & \textbf{0.02033} & \textbf{0.58301} & \textbf{0.01600} \\
Rambling-L & 0.70361& 0.03289 &0.68164 & 0.02609 & 0.65039 & 0.02057 & 0.60938 & 0.02068 \\
\bottomrule
\end{tabular}}
\end{minipage}
\end{table*}

\begin{table*}
\begin{minipage}{\textwidth}
\caption{The length and EOS rate of texts annotated by Video-LLaMA are measured on OpenVid-1M dataset to explore the influence of different perturbation magnitudes $\epsilon$. The best results are highlighted in \textbf{bold}.}
\vspace{-1em}
\label{tab:perturbation2}
\centering
\setlength\tabcolsep{8pt}{
\begin{tabular}{@{}l|cc|cc|cc|cc@{}}
\toprule
\multirow{2}{*}{Method} & \multicolumn{2}{c|}{$\epsilon$ = 2} & \multicolumn{2}{c|}{$\epsilon$ = 4} & \multicolumn{2}{c|}{$\epsilon$ = 8} & \multicolumn{2}{c}{$\epsilon$ = 16}\\ 

 & Length & EOS Rate(\%) & Length & EOS Rate(\%) & Length & EOS Rate(\%) & Length &EOS Rate(\%)  \\
\midrule 
 Original & 203.54 & 0.0 & 203.54 & 0.0 & 203.54 & 0.0 & 203.54 & 0.0\\
Noise & 209.81 & 0.0 & 222.06 & 0.0& 229.47 & 0.0 & 224.23 & 0.0  \\
Mute-S & 205.26& 0.0 & 205.73 & 0.0 & 131.32 & 5.0 & 11.58 & 7.0 \\
Mute-N & 217.00 & 0.0 & \textbf{192.73} & \textbf{10.0} & \textbf{48.61} & \textbf{79.0} & \textbf{0.00} & \textbf{100.0} \\
\bottomrule
\end{tabular}}
\end{minipage}
\end{table*}
\begin{table*}
\begin{minipage}{\textwidth}
\caption{Prompt transferability on non-annotation prompts: The length and EOS rate of texts generated by Video-LLaMA are measured on OpenVid-1M dataset. The prompt used during the attack phase is ``What is this video about?'' After the attack, we apply three different prompts—``Do you like this video?'', ``Who are you?'', and ``Where are you from?'' to generate text outputs for the videos. The best results are highlighted in \textbf{bold}.}
\vspace{-1em}
\label{tab:prompt 3}
\centering
\setlength\tabcolsep{17.5pt}{
\begin{tabular}{@{}l|cc|cc|cc@{}}
\toprule
\multirow{2}{*}{Method} & \multicolumn{2}{c|}{Do you like this video?} & \multicolumn{2}{c|}{Who are you?} & \multicolumn{2}{c}{Where are you from?}\\ 

 & Length & EOS Rate(\%) & Length & EOS Rate(\%) & Length & EOS Rate(\%)   \\
\midrule 
 Original & 245.33 & 0.0 & 158.58 & 0.0 & 153.59 & 0.0\\
Noise & 253.17 & 0.0 & 182.66 & 0.0 & 165.31 & 0.0\\
Mute-S & \textbf{36.05} & 8.0 & \textbf{22.23} & 4.0 & \textbf{36.50} & 4.0\\
Mute-N & 167.02 & \textbf{21.0} & 144.66 & \textbf{17.0} & 146.45 & \textbf{22.0}\\
\bottomrule
\end{tabular}}
\end{minipage}
\end{table*}

\begin{table*}
\begin{minipage}{\textwidth}
\caption{Ablation study on different inference parameters: The CLIP score and BLEU between texts annotated by Video-LLaMA and clean captions are measured on OpenVid-1M dataset. The parameters that are manually set are displayed below. DS means ``do\_sample'', NB means ``num\_beams'', T means temperature, and TP means ``top\_p''. The best results are highlighted in \textbf{bold}.}
\vspace{-1em}
\label{tab:para 1}
\centering
\setlength\tabcolsep{17.5pt}{
\begin{tabular}{@{}l|cc|cc|cc@{}}
\toprule
\multirow{2}{*}{Method} & \multicolumn{2}{c|}{DS=False} & \multicolumn{2}{c|}{DS=True,NB=2,T=1.0} & \multicolumn{2}{c}{DS=True,TP=0.8,T=0.8}\\ 
 & CLIP Score & BLEU & CLIP Score & BLEU & CLIP Score & BLEU   \\
\midrule 
 Original & 0.79248 & 0.05026 & 0.79346 & 0.04929 & 0.79443 & 0.05090\\
Noise & 0.79395 & 0.04794 & 0.79834 & 0.04818 &  0.79395 & 0.04959\\
Rambling-F & \textbf{0.58740} & \textbf{0.01684} & \textbf{0.58105} & \textbf{0.01663} & \textbf{0.57764} & \textbf{0.01595}\\
Rambling-L & 0.61328 & 0.02025 & 0.61719 & 0.02130 & 0.61621 & 0.02039\\
\bottomrule
\end{tabular}}
\end{minipage}
\end{table*}

\begin{table*}
\begin{minipage}{\textwidth}
\caption{Ablation study on different inference parameters: The length and EOS rate of texts annotated by Video-LLaMA are measured on OpenVid-1M dataset. The parameters that are manually set are displayed below. DS means ``do\_sample'', NB means ``num\_beams'', T means temperature, and TP means ``top\_p''. The best results are highlighted in \textbf{bold}.}
\vspace{-1em}
\label{tab:para 2}
\centering
\setlength\tabcolsep{17.7pt}{
\begin{tabular}{@{}l|cc|cc|cc@{}}
\toprule
\multirow{2}{*}{Method} & \multicolumn{2}{c|}{DS=False} & \multicolumn{2}{c|}{DS=True,NB=2,T=1.0} & \multicolumn{2}{c}{DS=True,TP=0.8,T=0.8}\\ 
 & Length & EOS Rate(\%) & Length & EOS Rate(\%) & Length & EOS Rate(\%)   \\
\midrule 
 Original & 221.61 & 0.0 & 213.19 & 0.0 & 197.12 & 0.0\\
Noise & 217.03 & 0.0 & 213.70 & 0.0 & 196.79 & 0.0\\
Mute-S & 14.76 & 8.0 & 27.63 & 2.0 & 21.82 & 8.0\\
Mute-N & \textbf{0.00} & \textbf{100.0} & \textbf{0.00} & \textbf{100.0} & \textbf{2.10} & \textbf{99.0}\\
\bottomrule
\end{tabular}}
\end{minipage}
\end{table*}

\subsection{Prompt Transferability of Mutes on Non-annotation Prompts}\label{app:prompt}
In the \cref{ablation}, we focus on prompts designed to annotate videos, such as ``What is this video about?''. Here, we extend this by examining the transferability of non-annotation prompts. During the attack phase, we maintain the same prompt-``What is this video about?'', as used in \cref{ablation}. And here, ``Do you like this video?'', ``Who are you?'', and ``Where are you from?'' are used to generate outputs. Since the datasets do not contain answers for these prompts, we focus on evaluating the performance of the Mutes method.

As shown in \cref{tab:prompt 3}, the performance of Mutes is evident in the decreased output length and higher EOS rate compared to the baselines. Notably, Mute-S excels in generating short sentences, as reflected in its great performance in output length, while Mute-N achieves a higher EOS rate, indicating its effectiveness in producing more null captions. Overall, this analysis highlights the prompt transferability of our Mutes method, even when applied to non-annotation prompts above.

\subsection{Ablation Study on Different Inference Parameters}\label{app:para}
In this section, we evaluate the robustness of our watermarking methods by conducting inferences using video-based LLMs with varying inference parameters. The maximum number of new tokens is set to 512. The parameters are displayed in \cref{tab:para 1} and \cref{tab:para 2}. As clearly shown, Ramblings and Mutes perform well under different inference parameters. Specifically, the CLIP score and BLEU decrease dramatically with Ramblings, while the EOS rate is high and the length is short with Mutes.

\subsection{Ablation Study on Text Perturbation}\label{app:text perturbation}
In this section, we fix the clean videos in the video-text pairs used to fine-tune text-to-video models and vary the texts to evaluate the impact of text perturbation. The results shown in \cref{tab:animate} and \cref{tab:animate ablation} indicate that the watermarks added to the videos play a significant role in reducing the performance of AnimateDiff. This is evident because the VQA\textsubscript{A} and VQA\textsubscript{T} remain relatively close to the original when only texts are altered and clean videos are fixed.

\subsection{Black-box Attack}\label{app:black}
Additionally, we perform black-box attacks to evaluate the transferability of our watermarking methods across different models. Adversarial videos are generated by Video-LLaMA based on the OpenVid-1M dataset and subsequently annotated by Video-Vicuna. In order to use Mutes to get shorter captions, we load two models simultaneously to annotate the adversarial videos for each optimization iteration, which is set as 200 (threshold is canceled), and select the minimum length as our results. As shown in \cref{tab:blackbox1} and \cref{tab:blackbox2}, both our Ramblings and Mutes demonstrate robust performance. The CLIP score, BLEU, and length show varying degrees of decline compared with baselines, indicating that our methods remain effective even without direct access to the victim model - Video-Vicuna.

\begin{table}
  \centering
  \caption{The VQA\textsubscript{A} and VQA\textsubscript{T} are evaluated for text-to-video models fine-tuned on different video-text pairs. All the videos are clean videos from OpenVid-1M. We change the texts to evaluate the influence of text perturbation when the videos are clean. Original means we use the texts that are annotated by Video-LLaMA on original videos. Rambling-F-Text-Only means we use the texts that are annotated by Video-LLaMA on videos protected by Rambling-F. The same applies to others. Specifically, the texts of Mute-N-Text-Only are null captions.}
  \setlength\tabcolsep{7pt}
  \begin{tabular}{@{}lcc@{}}
    \toprule
    Text & VQA\textsubscript{A} & VQA\textsubscript{T} \\
    \midrule
    Original & 60.430 & 51.302 \\
    Incorrect Text: Rambling-F-Text-Only & 60.717 & 45.724 \\
    Incorrect Text: Rambling-L-Text-Only & 62.218 & 50.322 \\
    Short Text: Mute-S-Text-Only & 60.717 & 45.425 \\
    Short Text: Mute-N-Text-Only & 62.377 & 52.008 \\
    \bottomrule
  \end{tabular}
  \label{tab:animate ablation}
\end{table}

\begin{table}
  \centering
  \caption{The CLIP score and BLEU of black-box attacks by Ramblings are evaluated. Here, we get the adversarial videos from Video-LLaMA and annotate these videos with Video-Vicuna. The dataset we used is OpenVid-1M. The best results are highlighted in \textbf{bold}.}
  \setlength\tabcolsep{9pt}
  \begin{tabular}{@{}l|cc|cc@{}}
    \toprule
     \multirow{2}{*}{Method} &\multicolumn{2}{c|}{CLIP Score} & \multicolumn{1}{c}{Video Caption}\\ 
      & RN50 & RN101 & BLEU \\
    \midrule 
    Original & 0.63818 & 0.75732 & 0.01969 \\
    Noise & 0.64453 & 0.76367 & 0.02275 \\
    Rambling-F & \textbf{0.43823} & \textbf{0.65674} & 0.01559 \\
    Rambling-L & 0.49438 & 0.66846 & \textbf{0.01441}\\
    \bottomrule
  \end{tabular}
  
  \label{tab:blackbox1}
\end{table}

\begin{table}
  \centering
  \caption{The length and EOS rate of black-box attacks by Mutes are evaluated. Here, we get the adversarial videos from Video-LLaMA and annotate these videos with Video-Vicuna. The dataset we used is OpenVid-1M. The best results are highlighted in \textbf{bold}.}
  \begin{tabular}{@{}l|cc@{}}
    \toprule
     Method & Length & EOS Rate(\%) \\
    \midrule 
    Original & 32.90 & 0.0 \\
    Noise & 38.27 & 0.0\\
    Mute-S &  \textbf{13.56} & 0.0\\
    Mute-N &  14.19 & 0.0\\
    \bottomrule
  \end{tabular}
  
  \label{tab:blackbox2}
\end{table}

\subsection{Explanation of Methods}\label{app:explain}
The reasons our methods disrupt annotation performance lie in the shifts within the feature space for Rambling-F and perturbations in token distribution for Rambling-L and Mutes. For Rambling-F, annotation performance decreases as the perturbation magnitude increases shown in \cref{tab:perturbation1}, accompanied by a growing shift distance in the feature space in \cref{tab:reason of method}. Additionally, the loss designs for Rambling-L and Mutes are specifically motivated by their impact on token distribution. Rambling-L shifts the token distribution away from the clean caption, whereas Mute-S and Mute-N focus on increasing the probability of the EOS token.
\begin{table}[htbp]
\centering
\caption{The feature variation of Rambling-F and the entropy of Rambling-L are evaluated under different perturbation magnitudes. The model and dataset we used are Video-LLaMA and OpenVid-1M respectively.}
\scriptsize
\vspace{-1.3em}
\setlength{\tabcolsep}{2.8mm}{
\begin{tabular}{l|cccc}
	\hline
	 Metric & $\epsilon$ = 2 & $\epsilon$ = 4 & $\epsilon$ = 8 & $\epsilon$ = 16 \\  \hline 
Video feature of Rambling-F& 0.4606 & 0.5252 & 0.6018 & 0.6629\\ 
LLM feature of Rambling-F& 0.5442 & 0.7552 & 0.9577 & 1.0961\\
Entropy of Rambling-L & 1.1167 & 1.8029 & 2.6964 & 4.0978\\
\hline
\end{tabular}}
\vspace{-1.5em}
\label{tab:reason of method}
\end{table}

\subsection{Adaptive Evaluation}\label{app:adaptive}
We evaluate the robustness of our methods under format conversion from avi to mkv in \cref{tab:mkv}. Specifically, we use the OpenCV package to save videos from avi to mkv using the FFV1 codec. The results below show that our methods can work well under format conversion from avi to mkv. 

We further conduct adaptive evaluation under compression. Given that compression is widespread in real-world applications, we consider a scenario where the video is resized from 224×224 to 112×112 for storage. To enhance adversarial performance, we incorporate an adaptive attack into our method by integrating downsampling and upsampling into the adversarial pipeline. As shown in \cref{tab:mkv}, our adaptive attack remains effective under video size compression. Additionally, we evaluate the robustness of our methods under noise removal using a mean filter. Similar to compression, we integrate the mean filter into the adversarial pipeline to enhance performance. The results demonstrate that our adaptive attack is also effective in mitigating the impact of the mean filter.
\begin{table}[htbp]
\centering
\caption{The adaptive evaluation of our methods. The model and dataset we used are Video-LLaMA and OpenVid-1M respectively.}
\scriptsize
\vspace{-1.3em}
\setlength{\tabcolsep}{0.3mm}{
\begin{tabular}{l|cc|cc|cc|cc}
	\hline
	 \multirow{2}{*}{Adaptability} & \multicolumn{2}{c|}{Rambling-F} & \multicolumn{2}{c|}{Rambling-L} & \multicolumn{2}{c|}{Mute-S} & \multicolumn{2}{c}{Mute-N} \\ 
     & CLIP & BLEU & CLIP & BLEU & Length & EOS Rate(\%) & Length & EOS Rate(\%)\\ \hline
     Original & 0.788 & 0.051 & 0.788 & 0.051 & 203.54 & 0.0 & 203.54 & 0.0\\
     AVI & 0.583 & 0.016 & 0.609 & 0.021 & 11.58 & 7.0 & 0.00 & 100.0\\
     MKV & 0.583 & 0.016 & 0.609 & 0.021 & 11.58 & 7.0 & 0.00 & 100.0\\
     Compress & 0.608 & 0.016 & 0.622 & 0.022 & 32.43 & 7.0 & 0.00 & 100.0\\
     Removal & 0.610 & 0.018 & 0.640 & 0.021 & 33.50 & 2.0 & 0.00 & 100.0\\

 \hline
\end{tabular}}
\vspace{-1.5em}
\label{tab:mkv}
\end{table}
\subsection{Optimization of Prompt Transferability}\label{optimization of prompt}To enhance the prompt transferability of Mute-N, we incorporate adversarial prompt training in CroPA~\cite{luo2024image} into the optimization for our attack, dubbed ``Mute-N2'' by adding adversarial perturbations both on prompt and video. The results in  \cref{tab:optimization of prompt} demonstrate that our Mute-N2 performs well across three prompts. Notably, only the single prompt ``What is this video about?'' is used for adversarial prompt training in Mute-N2.
\begin{table}[htbp]
\caption{The length and EOS rate of optimized method Mute-N2. The model and dataset we used are Video-LLaMA and OpenVid-1M respectively.}
\centering
\scriptsize
\vspace{-1.3em}
\setlength{\tabcolsep}{0.2mm}{
\begin{tabular}{l|cc|cc|cc}
	\hline
	 \multirow{3}{*}{Method} & \multicolumn{2}{c|}{\multirow{2}{*}{What is this video about?}} & \multicolumn{2}{c|}{\multirow{2}{*}{What happens in the video?}} & \multicolumn{2}{p{2cm}}{Can you describe the video in detail?} \\ 
     & Length & EOS Rate(\%) & Length & EOS Rate(\%) & Length & EOS Rate(\%) \\
     \hline 
     Mute-N2 & 0.00 & 100.0 & 2.67 & 99.0 & 86.53 & 68.0\\
 \hline
\end{tabular}}
\vspace{-1.6em}
\label{tab:optimization of prompt}
\end{table}
\subsection{Additional Evaluation on Perturbation}\label{addition evaluation} We conduct a human evaluation with five participants on ten video pairs. They rate the semantic similarity between clean and adversarial videos on a scale of 1 to 5, where 1 indicates poor semantic similarity and 5 indicates high semantic similarity. 
Finally, we use LPIPS~\cite{zhang2018perceptual} to measure the perceptual similarity. The results in ~\cref{tab:human evaluation} show that the adversarial videos are well-constrained.

\begin{table}[htbp]
\centering
\caption{Human and LPIPS evaluation between clean and adversarial videos. The model and dataset we used are Video-LLaMA and OpenVid-1M respectively.}
\scriptsize
\vspace{-1.3em}
\setlength{\tabcolsep}{3.7mm}{
\begin{tabular}{l|cccc}
	\hline
	 Metric & Rambling-F & Rambling-L & Mute-S & Mute-N \\  \hline 
Human & 4.10 & 4.32 & 4.36 & 4.22\\
LPIPS &  0.246 & 0.246 & 0.243 & 0.220\\ 

 \hline
\end{tabular}}
\vspace{-1.6em}
\label{tab:human evaluation}
\end{table}

\subsection{Visualization}\label{app:visualization}
\begin{figure*}
   
    \centering
    
    \includegraphics[width=0.85\linewidth]{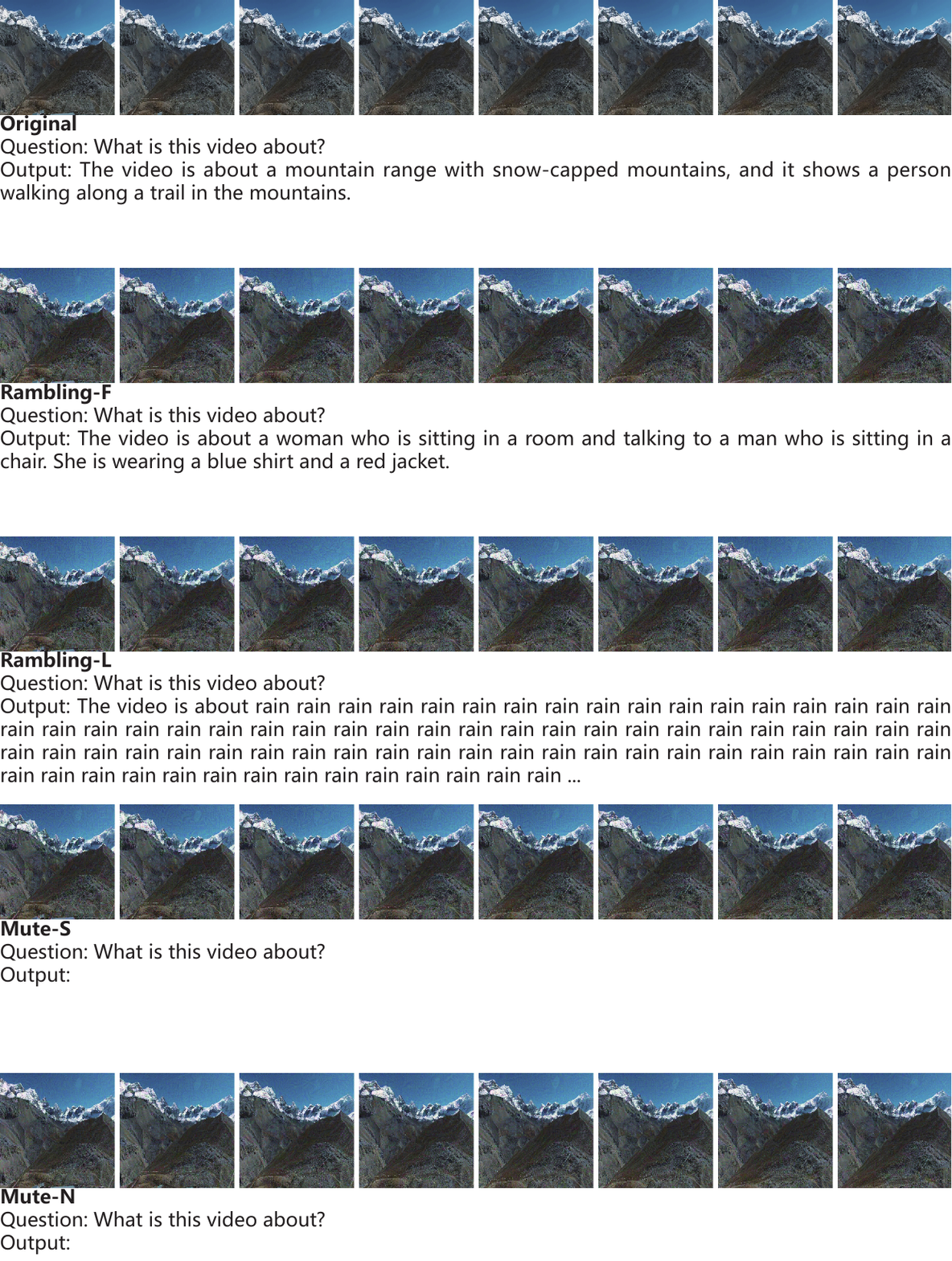}
    \caption{Visualization of watermarking methods against Video-ChatGPT.}
    \label{fig:visual7}
\end{figure*}

\begin{figure*}
   
    \centering
    
    \includegraphics[width=0.85\linewidth]{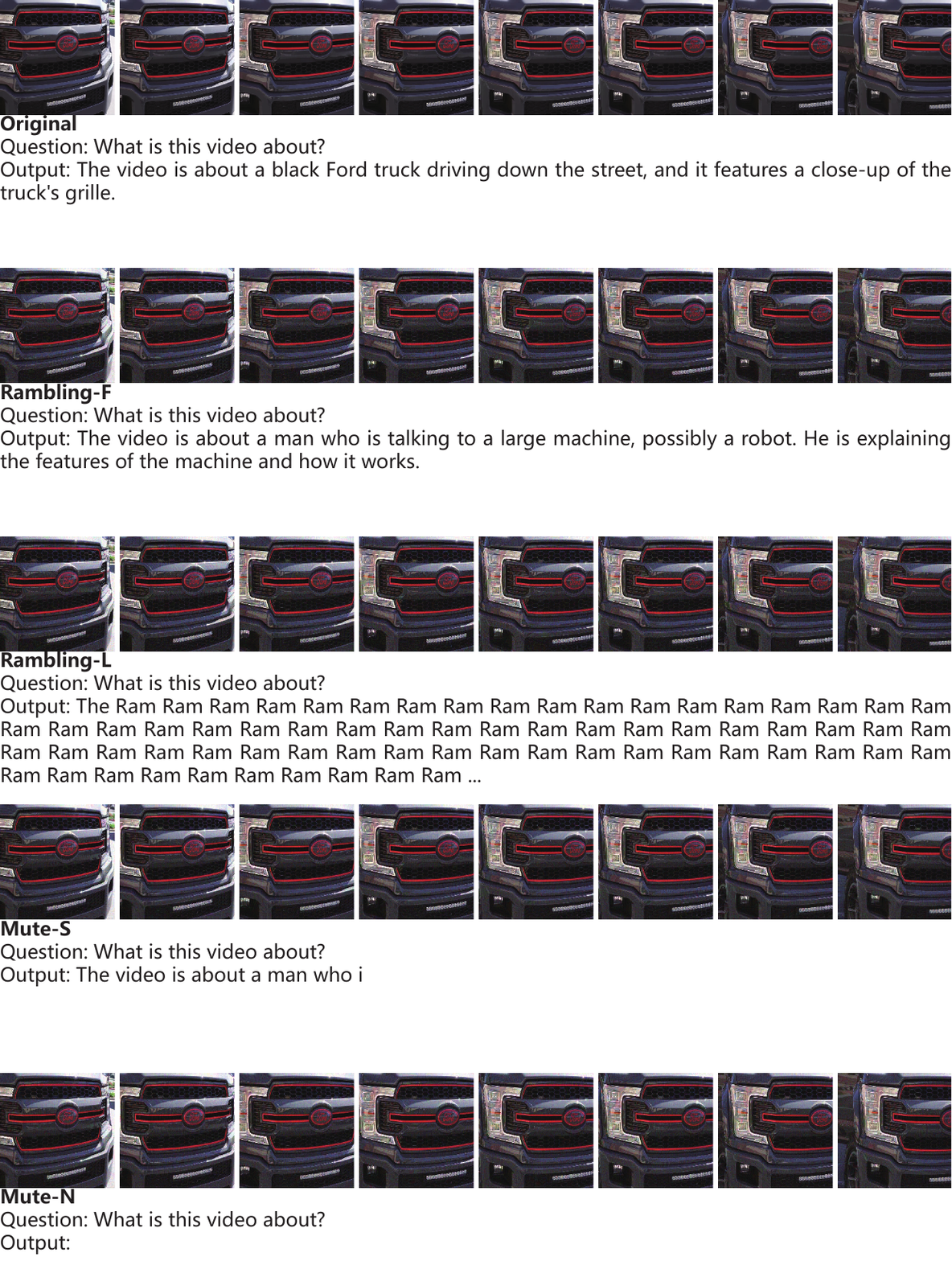}
    \caption{Visualization of watermarking methods against Video-ChatGPT.}
    \label{fig:visual8}
\end{figure*}

\begin{figure*}
   
    \centering
    
    \includegraphics[width=0.85\linewidth]{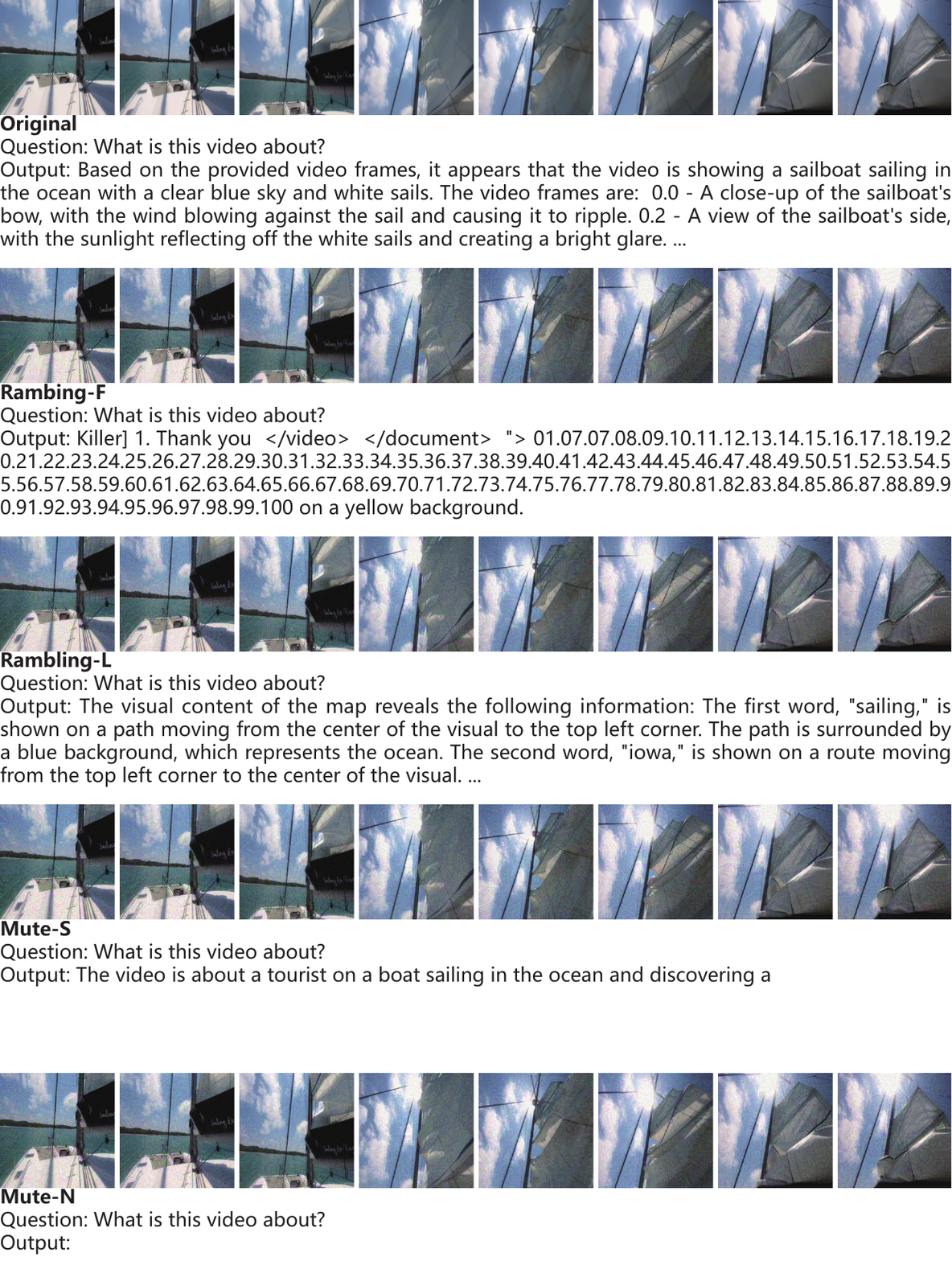}
    \caption{Visualization of watermarking methods against Video-LLaMA.}
    \label{fig:visual1}
\end{figure*}

\begin{figure*}
   
    \centering
    
    \includegraphics[width=0.85\linewidth]{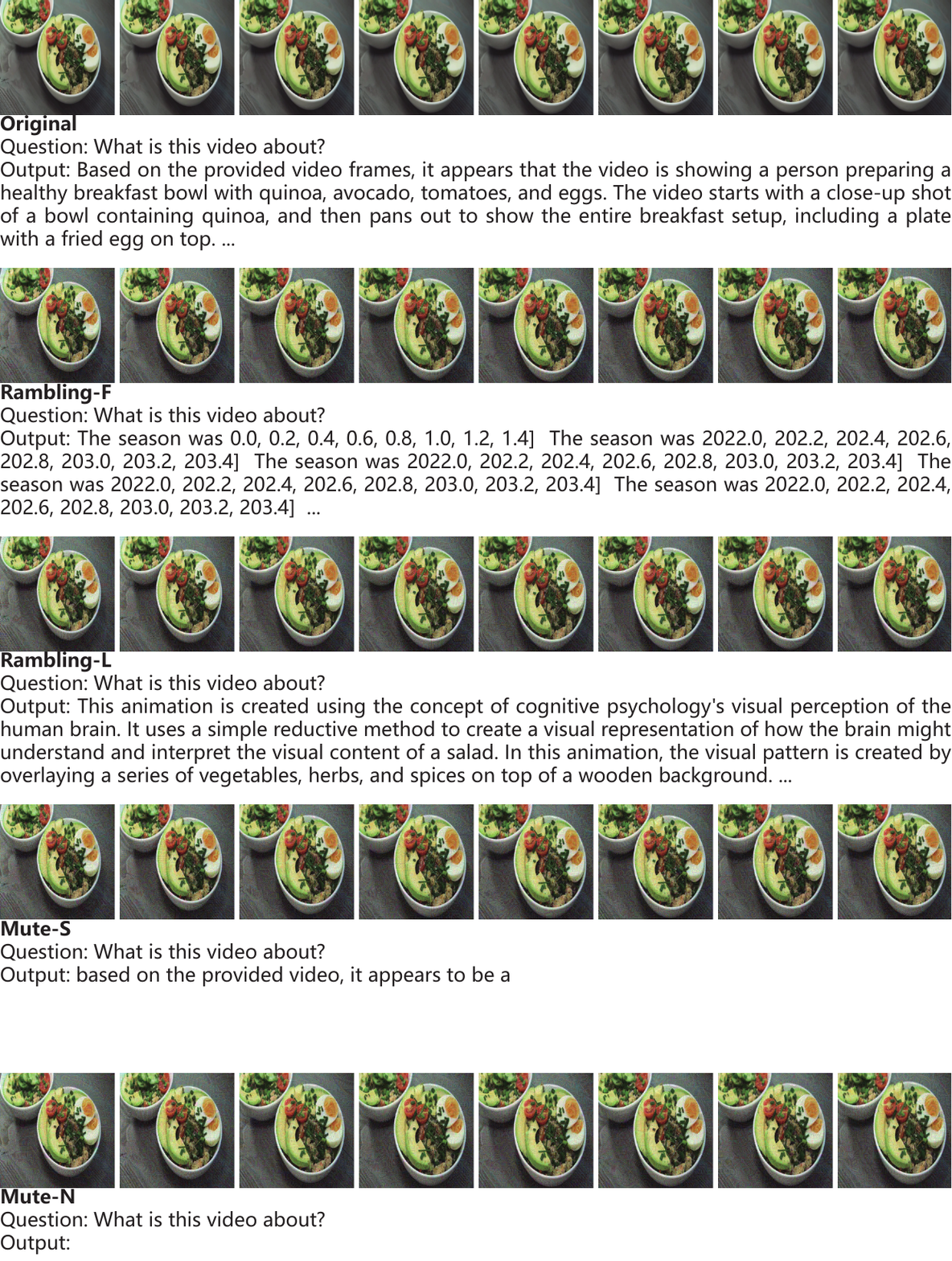}
    \caption{Visualization of watermarking methods against Video-LLaMA.}
    \label{fig:visual2}
\end{figure*}

\begin{figure*}
   
    \centering
    
    \includegraphics[width=0.85\linewidth]{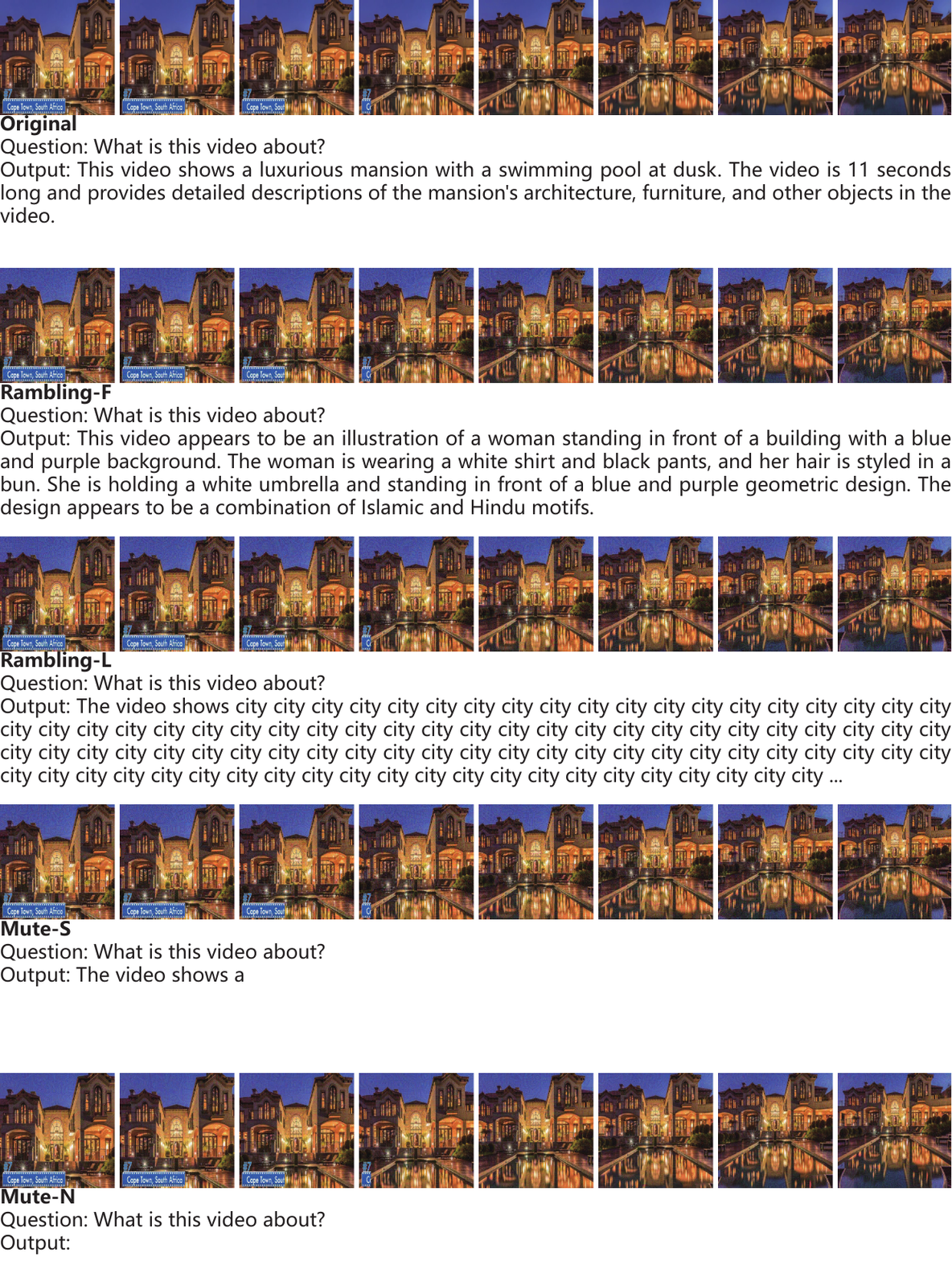}
    \caption{Visualization of watermarking methods against Video-Vicuna.}
    \label{fig:visual5}
\end{figure*}

\begin{figure*}
   
    \centering
    
    \includegraphics[width=0.85\linewidth]{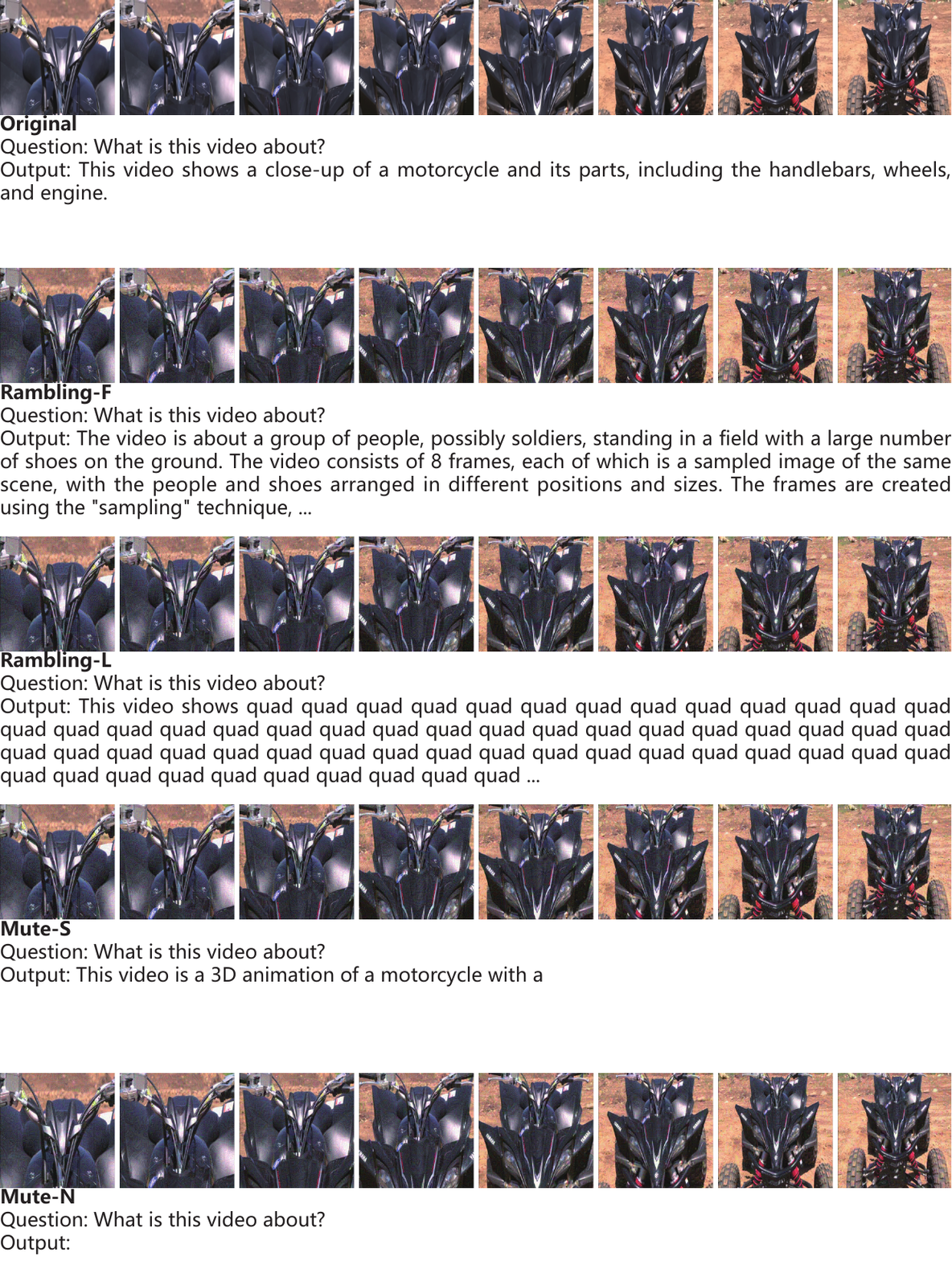}
    \caption{Visualization of watermarking methods against Video-Vicuna.}
    \label{fig:visual6}
\end{figure*}

The visualization of our watermarking methods is shown in \cref{fig:visual1,fig:visual2,fig:visual5,fig:visual6,fig:visual7,fig:visual8}. Our Ramblings and Mutes display great protective performance.

\clearpage

\end{document}